\pdfoutput=1

\documentclass[11pt]{article}

\usepackage[final]{acl}

\usepackage{times}
\usepackage{latexsym}
\usepackage{subcaption}
\usepackage{booktabs}
\usepackage{multirow}
\usepackage{tcolorbox}
\usepackage{amsmath}
\usepackage{float}
\usepackage{xcolor}

\usepackage[T1]{fontenc}
\usepackage[utf8]{inputenc}
\usepackage{CJKutf8}

\usepackage{microtype}

\usepackage{inconsolata}

\usepackage{graphicx}

\title{CS-Sum: A Benchmark for Code-Switching Dialogue Summarization and the Limits of Large Language Models}

\author{
  \textbf{Sathya Krishnan Suresh},
  \textbf{Tanmay Surana},
  \textbf{Lim Zhi Hao},
  \textbf{Eng Siong Chng}
\\
  Nanyang Technological University 
\\
  \small{\textbf{Correspondence:} \href{mailto:sathyakr001@e.ntu.edu.sg}{sathyakr001@e.ntu.edu.sg}}
}

\begin{document}
\maketitle
\begin{abstract}
Code-switching (CS) poses a significant challenge for Large Language Models (LLMs), yet its comprehensibility
remains underexplored in LLMs. We introduce \textbf{CS-Sum}, to evaluate the comprehensibility of 
CS by the LLMs through CS dialogue to English summarization. CS-Sum is the first benchmark for CS
dialogue summarization across Mandarin-English (EN-ZH), Tamil-English (EN-TA), and Malay-English (EN-MS), 
with 900-1300 human-annotated dialogues per language pair. Evaluating ten LLMs, including open and 
closed-source models, we analyze performance across few-shot, translate-summarize, and fine-tuning 
(LoRA, QLoRA on synthetic data) approaches. Our findings show that though the scores
on automated metrics are high, LLMs make subtle mistakes that alter the complete meaning
of the dialogue. To this end, we introduce 3 most common type of errors that LLMs make
when handling CS input. Error rates vary across CS pairs and LLMs, with some LLMs
showing more frequent errors on certain language pairs, underscoring the need for 
specialized training on code-switched data.
\end{abstract}

\section{Introduction}
Code-switching (CS) is the practice of alternating between two or more languages within a single conversation or utterance. 
Bilingual and multilingual speakers frequently engage in CS. This form of communication
has also become prevalent in social media and is the norm in multilingual societies. 
Large Language Models (LLMs) \cite{gpt4o,gemini,llama},
are trained on huge volumes of data majority of which is written in English. 
Despite the prevalence of CS in real-world communication, existing benchmarks 
\cite{huzaifah-etal-2024-evaluating,zhang-etal-2023-multilingual,yong-etal-2023-prompting}
fail to assess how well LLMs process CS dialogues.
Improved CS comprehension would enable LLMs to interpret multilingual prompts more effectively, enhancing accessibility for CS speakers \cite{ppl_prefer_code_mix_bots}. 

\begin{figure}
  \centering
  \includegraphics[width=0.8\columnwidth]{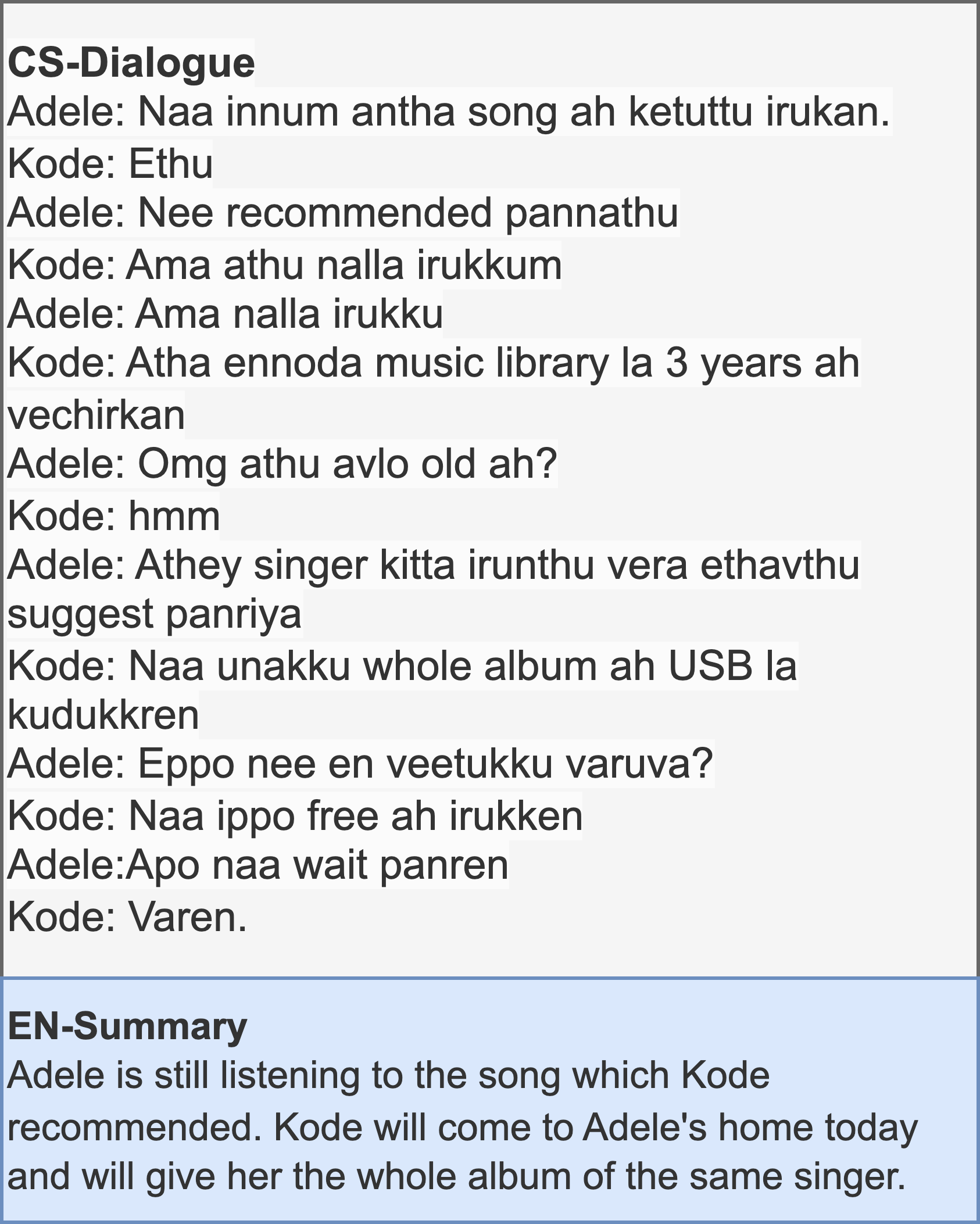}
  \caption{An instance from the benchmark}
  \label{fig:cs_example}
\end{figure}

Summarization is an ideal task to evaluate CS understanding, as it requires 
both grasping the dialogue’s main idea and generating a concise, coherent 
summary across languages. The only existing CS dialogue summarization dataset, 
Gupshup \cite{gupshup}, covers only one language pair (English-Hindi). 
Other CS benchmarks like LinCE \cite{lince} and GLUECos \cite{gluecos} 
focus on tasks such as language identification and POS-tagging, which assess 
word- or sentence-level understanding, whereas summarization requires discourse-level 
comprehension, a much harder test of CS ability.

To truly evaluate the ability of LLMs on CS, we propose the \textbf{CS-Sum} benchmark, 
which contains 900-1300 CS dialogue-summary pairs for three language pairs: 
Mandarin-English (EN-ZH), Tamil-English (EN-TA) and Malay-English (EN-MS). The CS dialogues were 
created by native speakers of the respective languages. Instead of generating
a dialogue from scratch, the speakers were asked to translate the English dialogues in test 
sets of DialogSum \cite{chen-etal-2021-dialogsum} and SAMSum \cite{gliwa-etal-2019-samsum} - 
two popular English dialogue summarization datasets - to 
CS dialogues in their respective languages. 

We evaluate 9 open source and 1 proprietary LLMs under few-shot,
translate-summarize, LoRA \cite{hu2022lora} and QLoRA \cite{qlora} on synthetic
data and provide a detailed analysis on the struggles of current SOTA LLMs(<=9B).
For LoRA and QLoRA fine-tuning, the training data was generated using Gemini-2-flash 
\cite{gemini}, as it is a large-scale LLM with a free-to-use API, making it a practical 
choice for our experiments.

\begin{table*}[t]
  \centering
  \begin{tabular}{cccccc}
    \hline
    \textbf{Language} & \textbf{M-Index} & \textbf{I-Index} & \textbf{Burstiness} & \textbf{Span Entropy} & \textbf{Memory}\\
    \hline
    EN-ZH & 0.40  & 0.36  & -0.79  & 0.13  & -0.08 \\
    EN-MS & 0.41  & 0.38  & -0.65  & 0.64  & -0.24\\
    EN-TA & 0.42  & 0.40  & -0.68  & 0.58  & -0.22\\
    \hline
  \end{tabular}
  \caption{
    CS metrics measured on the CS-Sum dataset
  }
  \label{tab:cs-metrics-cs-sum}
\end{table*}

Our qualitative analysis of the summaries generated by LLMs showed that tranditional 
summarization metrics like ROUGE \cite{lin-2004-rouge}, BERTScore \cite{zhang2019bertscore}
are not sufficient for measuring the quality of the summaries since the LLMs are capable 
of generating summaries that contain subtle errors that change the entire meaning of the 
CS dialogues. To this end, we identify the three most common errors (CSL, MST, SMA Section \ref{sec:error_analysis})
that the LLMs make when summarizing CS dialogues and suggest an LLM-driven 
approach to analyse their summaries with respect to the errors.

Our main contributions are:
\begin{itemize}
  \item We introduce CS-Sum, the first CS dialogue summarization benchmark for EN-ZH,
EN-TA and EN-MS.
  \item We conduct an in-depth evaluation of state-of-the-art LLMs, identifying critical failure 
patterns in their ability to process and summarize CS dialogues.
  \item We release CS-Sum, along with the full synthetic training dataset.
\end{itemize}

\section{Related Work}

While LLMs have demonstrated strong 
performance in many multilingual tasks, they still face significant 
challenges when dealing with CS.
Studies \cite{zhang-etal-2023-multilingual,yong-etal-2023-prompting} 
have shown that LLMs, including GPT-4 and GPT-3.5, struggle 
with generating code-switched text, often resulting in language 
collapse, where the model fails to mix languages properly, and in 
task-specific failures such as poor performance in summarization and 
machine translation for CS data. 
These issues arise because LLMs are typically trained on monolingual 
data and lack the capacity to handle the intricacies of CS discourse. 
Additionally \cite{huzaifah-etal-2024-evaluating}, their performance 
is inconsistent, particularly when 
translating low-resource language pairs, with models showing better 
results for high-resource languages but underperforming when 
confronted with languages that are underrepresented in training 
datasets. These findings highlight a big problem as people prefer 
to interact with LLMs in code-switch \cite{ppl_prefer_code_mix_bots}.

Recent benchmarks such as LinCE \cite{lince} and GLUECoS \cite{gluecos} 
have been instrumental in advancing the understanding of CS phenomena by 
focusing on tasks like LID, NER, and POS tagging. While these datasets have 
provided valuable insights, it's important to recognize that LLMs continue 
to encounter challenges when dealing with the complexities inherent in code-switched 
discourse. To truly evaluate the CS comprehension of LLMs, other benchmarks are 
necessary—ones that go beyond word-level tasks and assess deeper linguistic 
understanding. The first step in this direction is CS-Sum, a benchmark that 
evaluates LLMs on a more complex task, such as summarization, which requires a nuanced 
comprehension of code-switched text.

\section{CS-Sum Benchmark}

\begin{table}
  \centering
  \resizebox{\columnwidth}{!}{
    \begin{tabular}{cccc}
      \toprule
              & EN-ZH & EN-TA & EN-MS \\
      \midrule
      Number of instances               & 1320 & 1000 & 918 \\
      EN utterances                     & 1140 & 344 & 517 \\
      Lang utterances                   & 2079 & 983 & 265 \\
      EN as matrix language             & 4699 & 2340 & 2984 \\
      Lang as matrix language           & 4779 & 6790 & 1919 \\
      Avg. monolingual utterance length & 2.344 & 3.616 & 5.515 \\
      Avg. CS utterance length          & 5.816 & 9.642 & 13.885 \\
      \bottomrule
    \end{tabular}
  }
  \caption{CS statistics of CS-Sum. `Lang' refers to the language other than English}
  \label{tab:cs_stats}
\end{table}

\subsection{Dataset Construction}
The goal of building the CS-Sum benchmark, was to address the lack of a benchmark that evaluates 
the CS comprehension of LLMs. We selected Mandarin (zh), Bahasa Melayu (ms), and Tamil (ta) because 
these languages have large bilingual speaker communities that frequently code-switch with English. To efficiently create a 
diverse CS benchmark within resource constraints, we translated a subset of the combined test sets 
from DialogSum and SAMSum. We recruited 7 native speakers for each language to translate the 
English dialogues to CS dialogues in their respective languages. The translators were native 
speakers, all university students (bachelor’s or master’s level). 
The entire translation process lasted for about 5 months.

\subsection{Corpus Overview and Analysis}
\label{sec:corpus-analysis}

Table \ref{tab:cs_stats} presents key statistics of the 
CS-Sum benchmark across three language pairs. We observe 
that non-English utterances outnumber English ones in 
EN-ZH and EN-TA, while the reverse is true for EN-MS, 
suggesting stronger local language dominance in the 
former two. The matrix language distribution further 
highlights this variation: Tamil dominates as the matrix 
language in EN-TA, whereas EN-ZH and EN-MS are more 
balanced. This indicates differing CS behaviors—embedding 
in EN-TA versus alternation in EN-ZH and EN-MS.

Utterance lengths also vary significantly. EN-MS exhibits 
the longest average CS utterances (13.88 tokens), 
compared to 5.81 in EN-ZH, suggesting more complex 
span-level dependencies in Malay-English dialogues. The 
longer monolingual spans and varied matrix language 
roles imply that LLMs must manage different CS dynamics 
across language pairs.

\subsection{Quantifying Code-Switching}
\label{sec:cs-metrics}

We quantify the structural properties of code-switching in 
CS-Sum using five established metrics~\cite{cs_metrics}, 
reported in Table~\ref{tab:cs-metrics-cs-sum}. 
The \textbf{M-Index} measures the distributional balance between 
languages, with values near 1 indicating equal usage. 
The \textbf{I-Index} captures the token-level likelihood of 
language alternation. \textbf{Burstiness} characterizes the 
temporal irregularity of switching, where negative values imply 
regular alternation and positive values indicate clustering. 
\textbf{Span Entropy} quantifies the unpredictability in 
monolingual segment lengths, while \textbf{Memory} reflects the 
autocorrelation of consecutive spans—positive values suggest 
persistence, negative values suggest alternation.

Across the three CS pairs, CS-Sum exhibits moderate switching 
behavior. The M-Index values (0.40–0.42) suggest relatively 
balanced bilingual distributions, and I-Index scores (0.36–0.40) 
indicate frequent intra-sentential switching, most prominently 
in EN-TA. Burstiness values are consistently negative, 
reflecting structured rather than chaotic switching. Span 
Entropy is lowest for EN-ZH (0.13), indicating predictable 
switch boundaries, while EN-TA (0.58) and EN-MS (0.64) exhibit 
higher variability. Memory values are near zero or negative 
across all pairs, implying weak temporal consistency in span 
lengths, and highlighting the need for LLMs to dynamically 
track and adapt to switching boundaries during summarization.

\subsection{Gemini-Generated Training Data}
\label{sec:gemini-data}

\begin{figure*}[t]
  \centering
  % First row: en-zh and en-ta images
  \begin{subfigure}[b]{0.42\textwidth}
    \centering
    \includegraphics[width=0.9\textwidth]{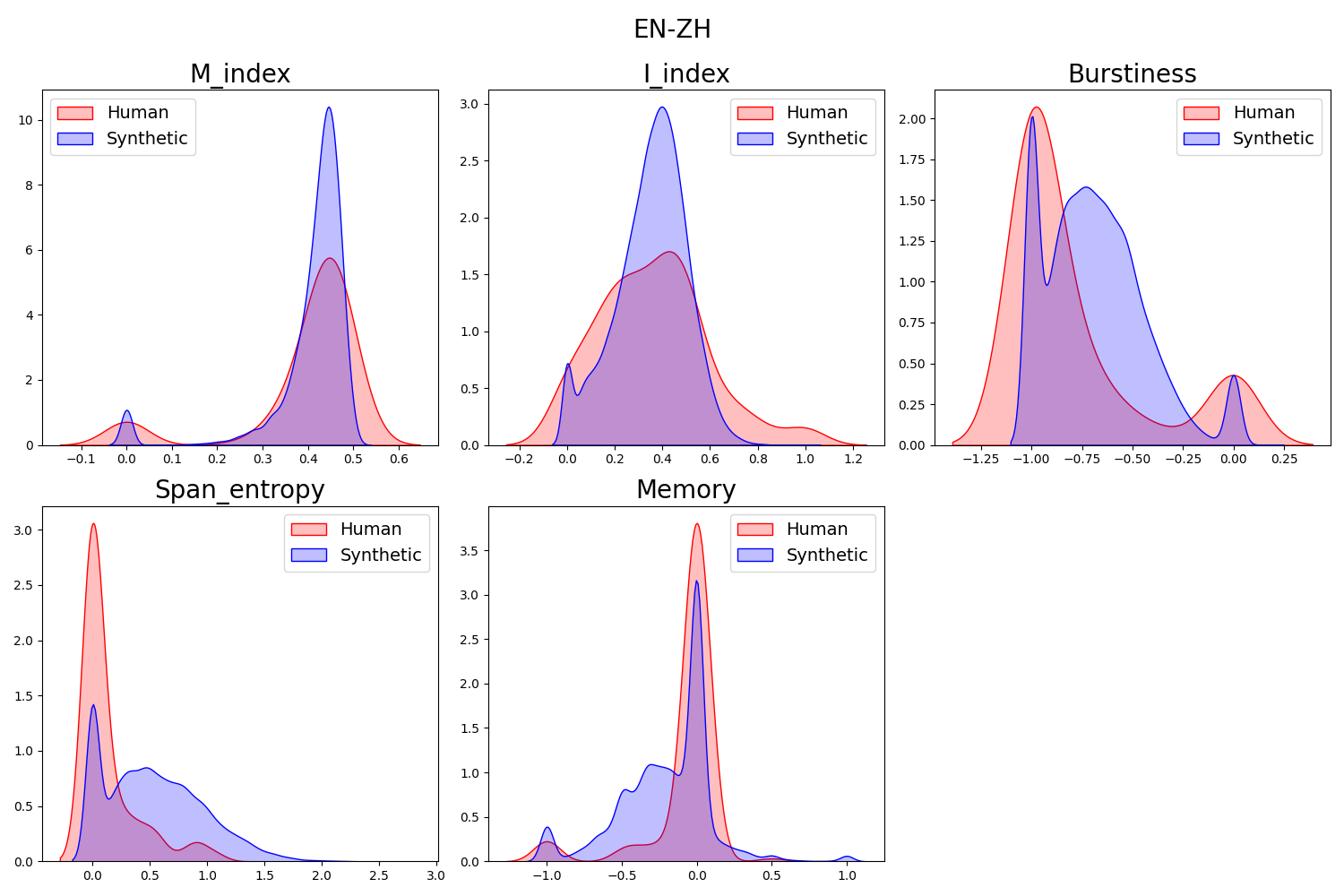}
    \caption{ZH Distribution}
    \label{fig:en_zh}
  \end{subfigure}%
  \hspace{0.05\textwidth}
  \begin{subfigure}[b]{0.42\textwidth}
    \centering
    \includegraphics[width=0.9\textwidth]{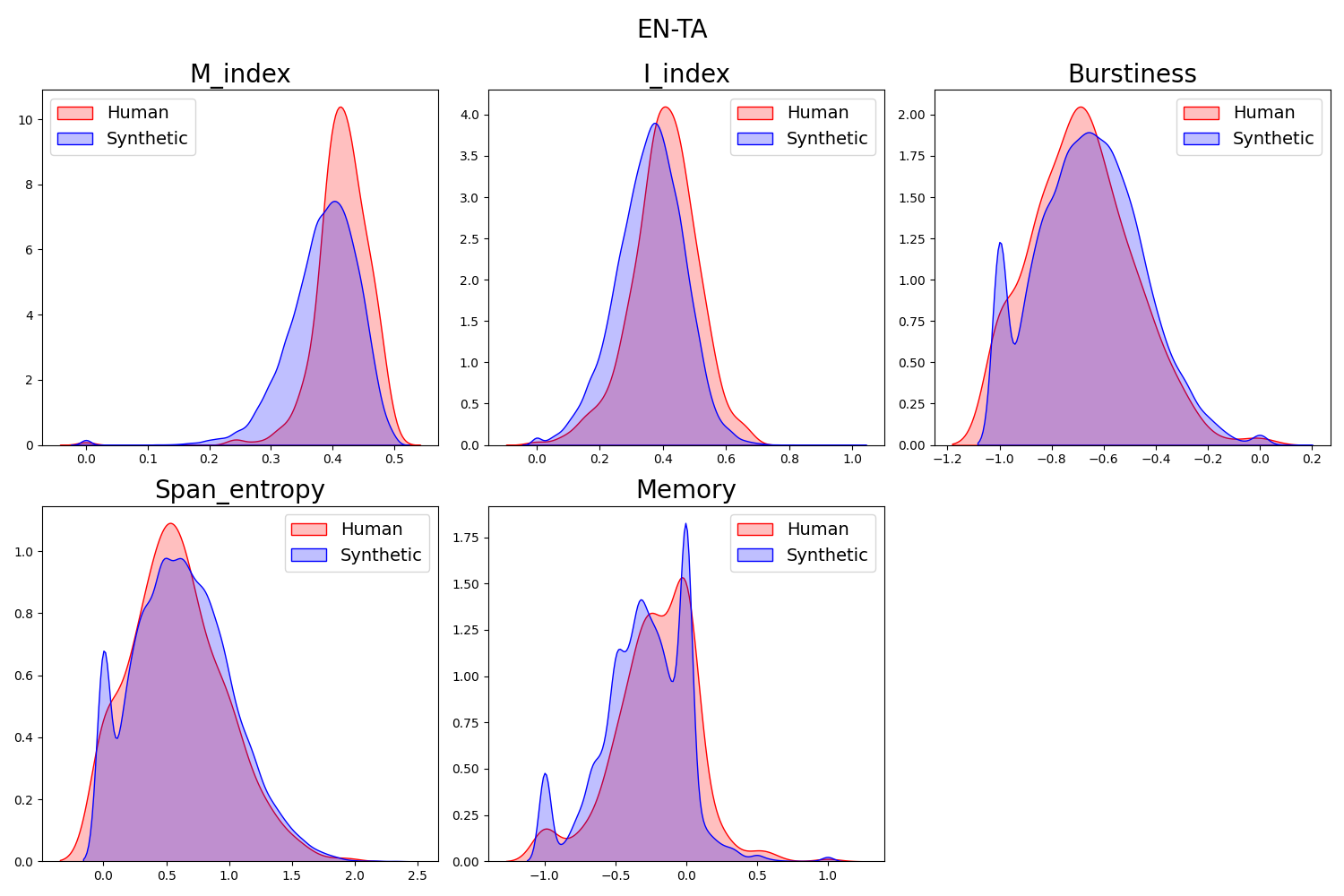}
    \caption{TA Distribution}
    \label{fig:en_ta}
  \end{subfigure}

  % Second row: en-ms image centered
  \vspace{0.02\textheight}
  \begin{subfigure}[b]{0.45\textwidth}
    \centering
    \includegraphics[width=0.9\textwidth]{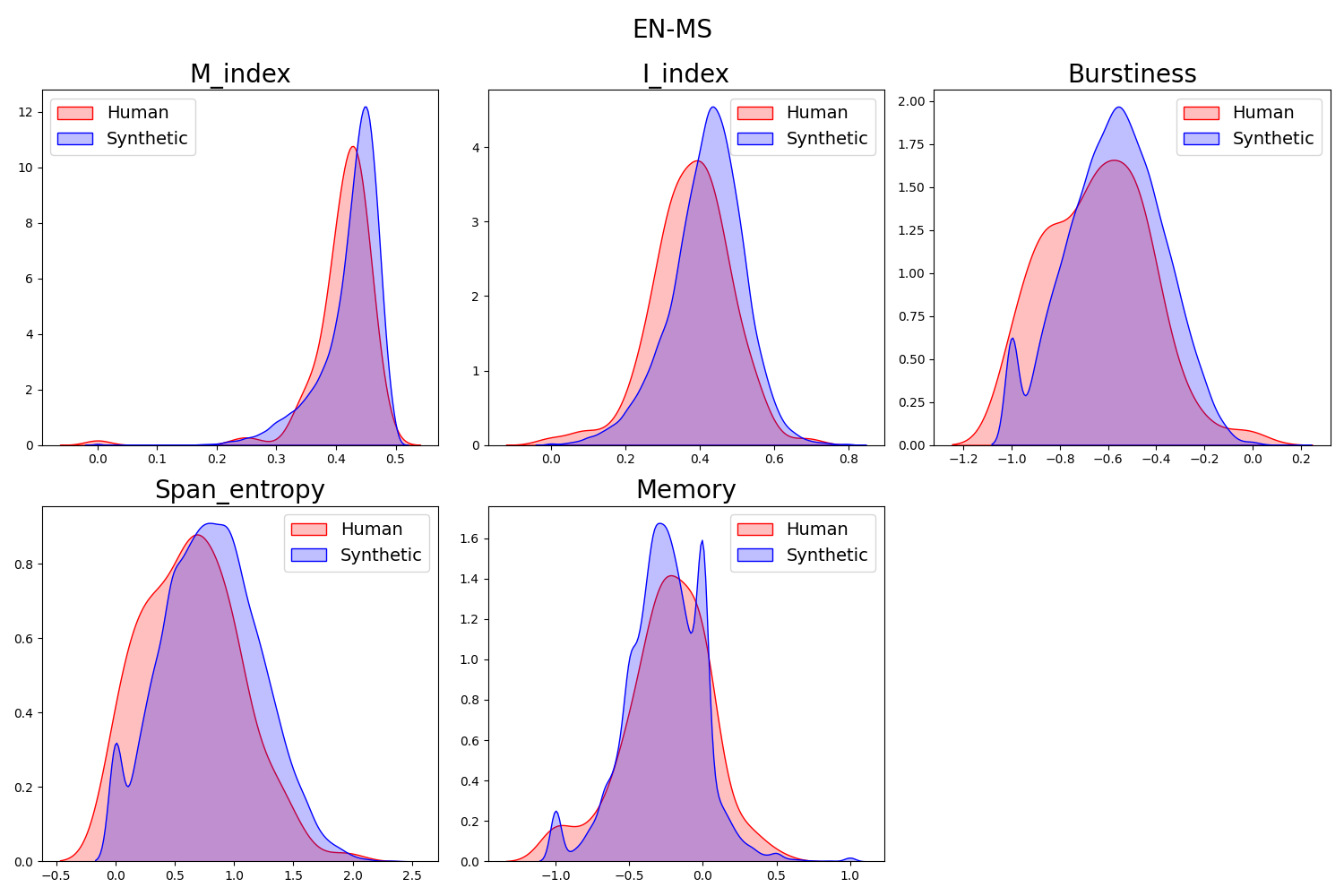}
    \caption{MS Distribution}
    \label{fig:en_ms}
  \end{subfigure}
\end{figure*}

\begin{table}
  \centering
  \begin{tabular}{ccc}
    \hline
    \textbf{CS pair} & \textbf{KL Div.} & \textbf{JS Div.} \\
    \hline
    EN-ZH & 2.4786 & 0.4918 \\ 
    EN-MS & 0.5481 & 0.1454 \\
    EN-TA & 0.5089 & 0.1017 \\
    \hline
  \end{tabular}
  \caption{Divergence between CS-Sum and CS-Sum-Syn}
  \label{tab:divergence}
\end{table}

To support fine-tuning for evaluating LLMs on CS-Sum, 
we construct a synthetic training set, \textbf{CS-Sum-Syn}, 
by converting 19,014 English dialogue-summary pairs from 
DialogSum and SAMSum into code-switched dialogues using 
Gemini-2\footnote{\url{https://ai.google.dev/}, gemini-2-flash-exp}~\cite{gemini}. 
This augmentation is not a contribution in itself, but a means 
to enable parameter-efficient adaptation (e.g., LoRA, QLoRA). To 
validate the distributional alignment of synthetic and human CS 
data, we compute CS metrics across both sets and visualize 
them for each language pair in Figures~\ref{fig:en_zh},~\ref{fig:en_ta}, 
and~\ref{fig:en_ms}.

We find that Gemini-generated data broadly approximates human 
CS patterns for EN-TA and EN-MS. In EN-TA, distributions 
over I-Index, burstiness, and span entropy align closely, 
suggesting that Gemini captures intra-sentential 
switching well. EN-MS also shows good overlap, though with
slightly more bursty and irregular switching in the synthetic 
set. In contrast, EN-ZH exhibits substantial divergence: 
synthetic dialogues show flatter memory and burstiness curves, 
as well as a long-tailed span entropy distribution, 
indicating reduced switching regularity and predictability. 
This mismatch is quantified in Table~\ref{tab:divergence}, 
where the KL and JS divergence for EN-ZH are 4--5× higher 
than for the other pairs. These discrepancies may affect the 
quality of fine-tuning and are revisited in 
Section~\ref{sec:error_analysis} during error diagnosis.

\section{LLM performance on CS-Sum}

\begin{table*}[t]
  \centering
  % \small
  \resizebox{\textwidth}{!}{
  \begin{tabular}{l|l|ccccc}
  \toprule
  \textbf{Model} & \textbf{Lang} & \textbf{ROUGE-L} & \textbf{BERTScore} & \textbf{SBERT-COSINE} & \textbf{JACCARD} & \textbf{METEOR} \\
  \midrule

  \multirow{3}{*}{\textbf{Gemma-2-2B}} 
  & EN-ZH & 0.2330 / \textcolor{red}{-4} & 0.8876 / \textcolor{red}{-0.35} & 0.7268 / \textcolor{red}{-1.40} & 0.1658 / \textcolor{red}{-7.64} & 0.2699 / \textcolor{red}{-0.9} \\
  & EN-TA & 0.2202 / \textcolor{red}{-7} & 0.8812 / \textcolor{red}{-0.38} & 0.6755 / \textcolor{red}{-0.85} & 0.1523 / \textcolor{red}{-9.19} & 0.2453 / \textcolor{red}{-5.23}\\
  & EN-MS & 0.2421 / \textcolor{red}{-7.18} & 0.8898 / \textcolor{red}{-0.98} & 0.7352 / \textcolor{red}{-4.35} & 0.1739 / \textcolor{red}{-13.52} & 0.2902 / \textcolor{red}{-12.43} \\
  \midrule

  \multirow{3}{*}{\textbf{Qwen2.5-3B}} 
  & EN-ZH & 0.2393 / \textcolor{green}{9.88} & 0.8824 / \textcolor{green}{0.93} & 0.7143 / \textcolor{green}{1.70} & 0.1643 / \textcolor{green}{5.82} & 0.2974 / \textcolor{green}{2.05} \\
  & EN-TA & 0.2227 / \textcolor{red}{-1.46} & 0.8805 / \textcolor{green}{0.44} & 0.6880 / \textcolor{red}{-2.35} & 0.1468 / \textcolor{red}{-0.13} & 0.2591 / \textcolor{red}{-4.82} \\
  & EN-MS & 0.2408 / \textcolor{green}{6.21} & 0.8868 / \textcolor{green}{0.60} & 0.7308 / \textcolor{red}{-2.25} & 0.1694 / \textcolor{green}{1.14} & 0.3100 / \textcolor{red}{-4.57} \\
  \midrule
  
  \multirow{3}{*}{\textbf{Qwen2.5-7B}} 
  & EN-ZH & 0.2482 / \textcolor{green}{27.10} & 0.8106 / \textcolor{green}{11.20} & 0.4927 / \textcolor{green}{54.12} & 0.0783 / \textcolor{green}{162.62} & 0.1118 / \textcolor{green}{188.89} \\
  & EN-TA & 0.2800 / \textcolor{red}{-5.11} & 0.8937 / \textcolor{red}{-0.55} & 0.7081 / \textcolor{red}{-2.14} & 0.1837 / \textcolor{red}{-9.62} & 0.2723 / \textcolor{red}{-9.20} \\
  & EN-MS & 0.3117 / \textcolor{red}{-1.65} & 0.9041 / \textcolor{red}{-0.46} & 0.7520 / \textcolor{red}{-0.67} & 0.2122 / \textcolor{red}{-8.37} & 0.3372 / \textcolor{red}{-10.26} \\
  \midrule

  \multirow{3}{*}{\textbf{SEALLM-7B}} 
  & EN-ZH & 0.2115 / \textcolor{green}{19.25} & 0.8122 / \textcolor{green}{9.15} & 0.4836 / \textcolor{green}{45.48} & 0.0834 / \textcolor{green}{89.27} & 0.1297 / \textcolor{green}{104.29} \\
  & EN-TA & 0.2270 / \textcolor{green}{0.99} & 0.8686 / \textcolor{green}{1.21} & 0.6470 / \textcolor{green}{2.53} & 0.1336 / \textcolor{green}{3.44} & 0.2287 / \textcolor{green}{1.55} \\
  & EN-MS & 0.2575 / \textcolor{green}{8.41} & 0.8801 / \textcolor{green}{1.28} & 0.6956 / \textcolor{green}{4.91} & 0.1633 / \textcolor{green}{9.04} & 0.2800 / \textcolor{green}{5.95} \\
  \midrule

  \multirow{3}{*}{\textbf{Mistral-7B}} 
  & EN-ZH & 0.2568 / \textcolor{red}{-3.10} & 0.8902 / \textcolor{red}{-0.42} & 0.7208 / \textcolor{red}{-0.34} & 0.1754 / \textcolor{red}{-8.01} & 0.3010 / \textcolor{red}{-7.82} \\
  & EN-TA & 0.2198 / \textcolor{red}{-2.15} & 0.8792 / \textcolor{green}{0.04} & 0.6847 / \textcolor{red}{-0.95} & 0.1470 / \textcolor{red}{-6.91} & 0.2657 / \textcolor{red}{-10.45} \\
  & EN-MS & 0.2421 / \textcolor{green}{1.59} & 0.8897 / \textcolor{red}{-0.46} & 0.7293 / \textcolor{red}{-2.15} & 0.1796 / \textcolor{red}{-11.49} & 0.3084 / \textcolor{red}{-8.40} \\
  \midrule

  \multirow{3}{*}{\textbf{Ministral-8B}} 
  & EN-ZH & 0.2560 / \textcolor{green}{4.23} & 0.8875 / \textcolor{green}{0.52} & 0.7037 / \textcolor{green}{1.06} & 0.1588 / \textcolor{green}{6.13} & 0.2517 / \textcolor{green}{1.15} \\
  & EN-TA & 0.2468 / \textcolor{red}{-5.61} & 0.8858 / \textcolor{red}{-0.18} & 0.6752 / \textcolor{red}{-2.95} & 0.1561 / \textcolor{red}{-3.66} & 0.2387 / \textcolor{red}{-8.53} \\
  & EN-MS & 0.2700 / \textcolor{green}{6.97} & 0.8917 / \textcolor{green}{0.37} & 0.7262 / \textcolor{green}{0.03} & 0.1794 / \textcolor{green}{2.93} & 0.2843 / \textcolor{red}{-0.09} \\
  \midrule

  \multirow{3}{*}{\textbf{LLaMA-3-8B}} 
  & EN-ZH & 0.2868 / \textcolor{red}{-2.29} & 0.8845 / \textcolor{green}{0.76} & 0.7090 / \textcolor{green}{4.09} & 0.1918 / \textcolor{red}{-6.99} & 0.2916 / \textcolor{green}{0.88} \\
  & EN-TA & 0.2453 / \textcolor{green}{0.47} & 0.8769 / \textcolor{green}{0.99} & 0.6707 / \textcolor{green}{3.28} & 0.1598 / \textcolor{green}{1.07} & 0.2570 / \textcolor{red}{-2.13} \\
  & EN-MS & 0.2599 / \textcolor{green}{8.06} & 0.8724 / \textcolor{green}{2.04} & 0.6538 / \textcolor{green}{11.21} & 0.1804 / \textcolor{green}{0.02} & 0.2833 / \textcolor{green}{5.50} \\
  \midrule

  \multirow{3}{*}{\textbf{Gemma-2-9B}} 
  & EN-ZH & 0.2995 / \textcolor{red}{-5.62} & 0.8987 / \textcolor{red}{-0.27} & 0.7603 / \textcolor{red}{-0.91} & 0.2054 / \textcolor{red}{-5.83} & 0.3246 / \textcolor{red}{-3.06} \\
  & EN-TA & 0.2761 / \textcolor{red}{-3.85} & 0.8911 / \textcolor{green}{0.57} & 0.7203 / \textcolor{green}{4.42} & 0.1832 / \textcolor{green}{4.63} & 0.3107 / \textcolor{green}{1.25} \\
  & EN-MS & 0.3009 / \textcolor{red}{-3.00} & 0.8975 / \textcolor{green}{0.01} & 0.7573 / \textcolor{red}{-0.94} & 0.2096 / \textcolor{red}{-3.96} & 0.3279 / \textcolor{red}{-0.07} \\
  \midrule

  \multirow{3}{*}{\textbf{SEA-LION-Gemma-2-9B}} 
  & EN-ZH & 0.2799 / \textcolor{red}{-3.96} & 0.8968 / \textcolor{red}{-0.34} & 0.7542 / \textcolor{red}{-0.41} & 0.1820 / \textcolor{red}{-4.29} & 0.2944 / \textcolor{red}{-4.46} \\
  & EN-TA & 0.2703 / \textcolor{green}{0.26} & 0.8898 / \textcolor{green}{0.48} & 0.7138 / \textcolor{green}{3.79} & 0.1727 / \textcolor{red}{-0.12} & 0.2884 / \textcolor{red}{-2.38} \\
  & EN-MS & 0.2860 / \textcolor{red}{-1.56} & 0.8963 / \textcolor{red}{-0.12} & 0.7404 / \textcolor{green}{1.21} & 0.1837 / \textcolor{red}{-4.16} & 0.2983 / \textcolor{red}{-4.75} \\
  \midrule

  \multirow{3}{*}{\textbf{GPT-4o}} 
  & EN-ZH & 0.2965 / \textcolor{red}{-1.03} & 0.8816 / \textcolor{green}{2.21} & 0.7082 / \textcolor{green}{9.63} & 0.1792 / \textcolor{green}{11.66} & 0.2853 / \textcolor{green}{16.52} \\
  & EN-TA & 0.3157 / \textcolor{red}{-4.92} & 0.8981 / \textcolor{green}{0.39} & 0.7570 / \textcolor{green}{2.19} & 0.2041 / \textcolor{red}{-0.55} & 0.3419 / \textcolor{green}{0.75} \\
  & EN-MS & 0.3126 / \textcolor{red}{-1.70} & 0.9037 / \textcolor{red}{-0.15} & 0.7750 / \textcolor{green}{0.12} & 0.2045 / \textcolor{green}{0.33} & 0.3365 / \textcolor{green}{1.09} \\

  \bottomrule
  \end{tabular}
  }
  \caption{Few-Shot Performance/Translate-Summarize \% improvement on CS-Sum}
  \label{tab:few_shot_all_ts}
\end{table*}

In this section, we present the performance of ten LLMs—nine 
open-source\footnote{Instruction-tuned and latest versions} and one 
proprietary—on CS-Sum across four settings: Few-Shot, 
Translate-Summarize, LoRA, and QLoRA. In LoRA and QLoRA, 
open-source models are fine-tuned on CS-Sum-Syn. The evaluated 
open-source LLMs include \textsc{LLaMA-3-8B} \cite{llama}, 
\textsc{Mistral-7B} \cite{Jiang2023Mistral7}, 
\textsc{Ministral-8B}\footnote{\url{https://huggingface.co/mistralai/Ministral-8B-Instruct-2410}}, 
\textsc{Gemma-2-2B} and \textsc{Gemma-2-9B} 
\cite{team2024gemma}, \textsc{Qwen2.5-2B} and 
\textsc{Qwen2.5-7B} \cite{yang2024qwen2}, 
\textsc{SEA-LION-Gemma-2-9B}\footnote{\url{https://sea-lion.ai/}}, 
and \textsc{SEALLM-7B} \cite{nguyen-etal-2024-seallms}. 
The proprietary model used is \textsc{GPT-4o} \cite{gpt4o}. 
These models were selected based on their multilingual 
pretraining exposure and computational feasibility, 
with sizes ranging from 2B to 9B parameters. Models like 
\textsc{LLaMA-3-8B}, \textsc{Mistral-7B}, and \textsc{Gemma-2-9B} have 
shown strong generalization across NLP tasks, while 
\textsc{Qwen2.5-7B}, \textsc{SEA-LION-Gemma-2-9B}, and 
\textsc{SEALLM-7B} were chosen for their targeted 
multilingual capabilities, especially in languages like 
Bahasa Melayu, Tamil, and Mandarin. \textsc{GPT-4o} serves 
as a strong proprietary baseline due to its 
SOTA performance in cross-lingual comprehension.

By evaluating these models across different parameter scales 
and varying degrees of multilingual exposure, we 
investigate how model size and pretraining diversity affect 
performance. LoRA and QLoRA fine-tuning on CS-Sum-Syn result 
in higher scores on automated metrics, suggesting improved 
alignment with reference summaries. However, it remains 
unclear whether these gains reflect actual comprehension of 
code-switched dialogue or simply the ability to replicate the 
distributional patterns of Gemini-generated synthetic data. 
To explore this, we conduct an in-depth error analysis in 
Section~\ref{sec:error_analysis}, which reveals that 
fine-tuning on synthetic data does not improve—and often 
degrades—the model’s ability to accurately interpret and 
summarize code-switched content.

\subsection{Result Analysis}

In this subsection, we present the results of the LLMs' performance
on the CS-Sum in Few-Shot, Translate-Summarize, LoRA and QLoRA settings.
Appendix \ref{app:experiment_settings} provides an explanation on the different
settings. We use standard summarization metrics like ROUGE \cite{lin-2004-rouge},
BERTScore \cite{zhang2019bertscore}, SBERT-Cosine \cite{sbert}, 
Jaccard and METEOR \cite{banerjee-lavie-2005-meteor}. This comprehensive 
set of metrics evaluate both word-level matching and semantic similarity.

\begin{figure}[t]
  \centering
  \includegraphics[width=\linewidth]{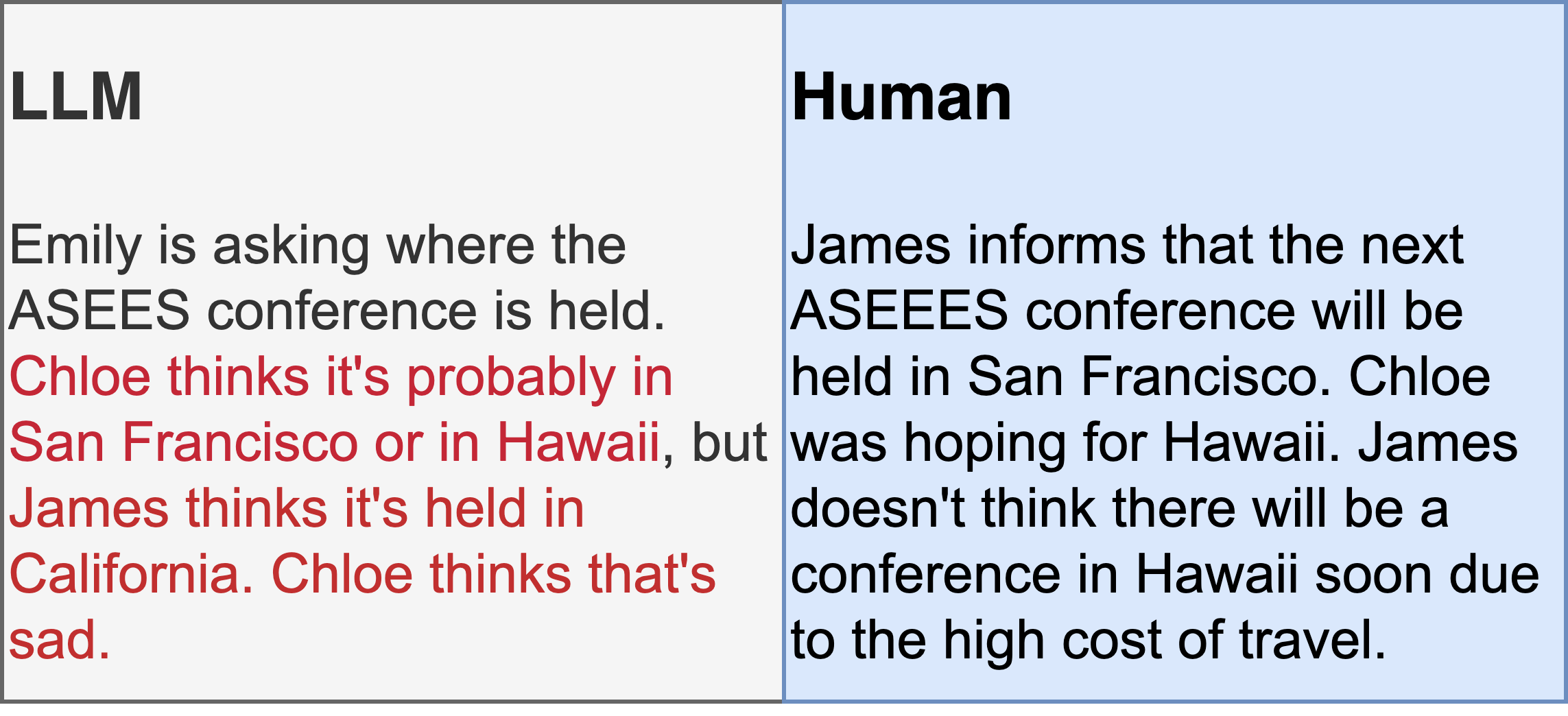}  % Replace with your image file
  \caption{Wrong summary with high BERTScore \(0.903\)}
  \label{fig:error_case}
\end{figure}

The few-shot performance of LLMs across the three CS 
language pairs is reported in Table~\ref{tab:few_shot_all_ts}. 
While semantic similarity metrics such as BERTScore and 
SBERT-Cosine report high values, these can be misleading. 
As illustrated in Figure~\ref{fig:error_case}, the 
summaries often fail to accurately capture the core content 
or intent of the original dialogue, despite appearing 
semantically plausible. This reflects a common failure mode 
of LLMs in CS summarization: generating fluent, high-overlap 
summaries that omit or distort key factual elements. In this 
example, Gemma-2-9B misrepresents who confirms the location 
of the conference and fabricates speaker opinions, despite 
achieving a high semantic similarity score. ROUGE-L scores 
further corroborate this gap, with LLMs scoring nearly 
50\% lower than the best-performing models on the monolingual 
DialogSum\footnote{\url{https://paperswithcode.com/sota/text-summarization-on-dialogsum}} and 
SAMSum\footnote{\url{https://paperswithcode.com/sota/text-summarization-on-samsum-corpus}} benchmarks.

From Table~\ref{tab:few_shot_all_ts}, we observe that GPT-4o 
is the best-performing model overall. However, due to the 
lack of transparency in its training and architecture, we 
cannot provide deeper analysis on its performance 
characteristics. Among open-source models, Gemma-2-9B 
consistently outperforms others across most metrics and 
language pairs. This aligns with findings from the Gemma-2 
paper~\cite{team2024gemma}, which attributes its strong 
multilingual performance to extensive exposure to diverse 
languages and an optimized tokenizer. Interestingly, 
SEA-Lion-Gemma-2-9B lags slightly behind, despite its regional 
specialization in Southeast Asian languages, suggesting that 
broader multilingual exposure may be more beneficial than 
region-specific pretraining for CS comprehension.

One notable observation is the underperformance of most 
models on the EN-ZH pair, which is surprising given that all 
models were exposed to Mandarin during pretraining. Manual 
inspection reveals that some LLMs generate summaries 
partially or entirely in Chinese, despite being explicitly 
prompted to produce English outputs. This mismatch leads to 
lower scores on automated metrics, particularly those that 
penalize non-English outputs. However, as discussed in our 
fine-grained evaluation (Section~\ref{sec:error_analysis}), 
these summaries often contain fewer semantic or structural 
errors than those in EN-TA or EN-MS, indicating that the 
low metric scores may not reflect actual comprehension quality.

We also find that smaller models, such as Gemma-2-2B and 
Qwen2.5-3B, perform significantly worse than their larger 
counterparts, reinforcing the impact of model scale on CS 
summarization. Surprisingly, SEALLM-7B performs comparably to 
these smaller models, despite being explicitly trained on 
Southeast Asian languages. Qualitative analysis of its 
outputs reveals several failure modes in the few-shot 
setting: \textbf{(a)} copying utterances directly from the 
dialogue, \textbf{(b)} generating descriptive paraphrases 
rather than true summaries, and \textbf{(c)} producing 
outputs in the wrong language. These behaviors suggest that 
SEALLM struggles with in-context learning and generalization, 
even within its intended linguistic domain.

\paragraph{Translate-Summarize \% Improvement}
Table~\ref{tab:few_shot_all_ts} reports percentage 
improvements under the Translate-Summarize setting. Contrary 
to expectations, most models show little to no improvement, 
and in many cases, performance declines across all metrics. 
This suggests that translation introduces additional 
challenges, likely due to LLMs’ inability to preserve the 
structural and discourse-level nuances of code-switched 
input. Notably, Qwen2.5-7B and SEALLM-7B improve 
significantly for EN-ZH (27.1\% and 19.25\% in ROUGE-L), 
reflecting benefits from Mandarin or SEA-specific 
pretraining. However, similar gains are absent for EN-TA and 
EN-MS, indicating that CS translation remains unreliable 
outside of high-resource language pairs.

\begin{table*}[t]
  \centering
  % \small
  \resizebox{\textwidth}{!}{
  \begin{tabular}{l|l|ccccc}
  \toprule
  \textbf{Model} & \textbf{Lang} & \textbf{ROUGE-L} & \textbf{BERTScore} & \textbf{SBERT-COSINE} & \textbf{JACCARD} & \textbf{METEOR} \\
  \midrule

  \multirow{3}{*}{\textbf{SEALLM-7B}} 
  & EN-ZH & \textcolor{green}{50.45} / \textcolor{green}{51.94} & \textcolor{green}{11.09} / \textcolor{green}{11.21} & \textcolor{green}{60.22} / \textcolor{green}{60.69} & \textcolor{green}{170.47} / \textcolor{green}{160.94} & \textcolor{green}{163.13} / \textcolor{green}{142.51} \\
  & EN-TA & \textcolor{green}{47.99} / \textcolor{green}{43.82} & \textcolor{green}{4.49} / \textcolor{green}{4.52} & \textcolor{green}{19.84} / \textcolor{green}{19.86} & \textcolor{green}{75.18} / \textcolor{green}{69.14} & \textcolor{green}{47.14} / \textcolor{green}{35.42} \\
  & EN-MS & \textcolor{green}{34.20} / \textcolor{green}{17.03} & \textcolor{green}{3.29} / \textcolor{green}{2.53} & \textcolor{green}{12.01} / \textcolor{green}{27.78} & \textcolor{green}{47.49} / \textcolor{green}{4.22} & \textcolor{green}{13.08} / \textcolor{green}{27.78} \\
  \midrule
  
  \multirow{3}{*}{\textbf{LLaMA-3-8B}} 
  & EN-ZH & \textcolor{green}{23.51} / \textcolor{green}{25.21} & \textcolor{green}{3.14} / \textcolor{green}{3.12} & \textcolor{green}{12.88} / \textcolor{green}{12.79} & \textcolor{green}{29.85} / \textcolor{green}{34.26} & \textcolor{green}{19.70} / \textcolor{green}{29.68} \\
  & EN-TA & \textcolor{green}{51.45} / \textcolor{green}{47.07} & \textcolor{green}{4.32} / \textcolor{green}{4.00} & \textcolor{green}{18.91} / \textcolor{green}{18.28} & \textcolor{green}{66.19} / \textcolor{green}{60.27} & \textcolor{green}{39.72} / \textcolor{green}{42.35} \\
  & EN-MS & \textcolor{green}{46.81} / \textcolor{green}{21.73} & \textcolor{green}{4.85} / \textcolor{green}{3.27} & \textcolor{green}{22.19} / \textcolor{green}{16.94} & \textcolor{green}{50.56} / \textcolor{green}{23.01} & \textcolor{green}{39.34} / \textcolor{green}{17.83} \\
  \midrule

  \multirow{3}{*}{\textbf{Gemma-2-9B}} 
  & EN-ZH & \textcolor{red}{-1.45} / \textcolor{green}{10.66} & \textcolor{green}{0.12} / \textcolor{green}{0.82} & \textcolor{green}{4.10} / \textcolor{green}{2.32} & \textcolor{green}{8.56} / \textcolor{green}{11.61} & \textcolor{green}{1.68} / \textcolor{green}{7.95} \\
  & EN-TA & \textcolor{green}{6.54} / \textcolor{green}{19.37} & \textcolor{green}{0.63} / \textcolor{green}{1.56} & \textcolor{green}{7.14} / \textcolor{green}{9.12} & \textcolor{green}{18.12} / \textcolor{green}{27.89} & \textcolor{green}{7.28} / \textcolor{green}{13.01} \\
  & EN-MS & \textcolor{green}{7.81} / \textcolor{red}{-6.27} & \textcolor{green}{0.71} / \textcolor{red}{-0.61} & \textcolor{green}{3.60} / \textcolor{red}{-1.41} & \textcolor{green}{8.48} / \textcolor{red}{-3.45} & \textcolor{green}{5.31} / \textcolor{red}{-7.06} \\
  \midrule

  \multirow{3}{*}{\textbf{SEA-LION-Gemma-2-9B}} 
  & EN-ZH & \textcolor{red}{-22.93} / \textcolor{red}{-6.64} & \textcolor{red}{-3.10} / \textcolor{red}{-0.71} & \textcolor{red}{-3.27} / \textcolor{green}{1.73} & \textcolor{red}{-4.13} / \textcolor{green}{8.67} & \textcolor{green}{6.29} / \textcolor{green}{14.39} \\
  & EN-TA & \textcolor{red}{-21.48} / \textcolor{green}{3.59} & \textcolor{red}{-2.61} / \textcolor{green}{0.24} & \textcolor{green}{1.74} / \textcolor{green}{7.08} & \textcolor{red}{-3.49} / \textcolor{green}{20.38} & \textcolor{green}{7.38} / \textcolor{green}{17.21} \\
  & EN-MS & \textcolor{red}{-21.98} / \textcolor{red}{-8.08} & \textcolor{red}{-2.95} / \textcolor{red}{-0.99} & \textcolor{red}{-3.53} / \textcolor{red}{-0.77} & \textcolor{red}{-8.16} / \textcolor{green}{3.18} & \textcolor{green}{9.09} / \textcolor{green}{3.03} \\

  \bottomrule
  \end{tabular}
  }
  \caption{LoRA / QLoRA \% Improvement over Few-Shot on CS-Sum}
  \label{tab:lora_qlora_imp}
\end{table*}

\paragraph{LoRA and QLoRA Results}
We fine-tune LLMs on the synthetic CS-Sum-Syn dataset using 
LoRA and QLoRA to evaluate their ability to handle 
CS summarization beyond just metric gains. As shown in 
Table~\ref{tab:lora_qlora_imp} \footnote{Due to space constraints, rest of the scores are presented in Table \ref{tab:lora_qlora_imp_2}}, 
EN-ZH sees the largest 
improvements (e.g., SEALLM-7B and Qwen2.5-7B exceed 200\% in 
METEOR), EN-TA shows consistent gains, while EN-MS exhibits 
mixed results across metrics.

These improvements are surprising given the distributional 
mismatch between CS-Sum and CS-Sum-Syn (Figure~\ref{fig:en_zh}, 
Table~\ref{tab:divergence}). A qualitative review reveals that 
models often produce summaries with subtle semantic errors, 
indicating they may have learned surface-level summary 
structure rather than true code-switching comprehension. 
This motivates the deeper analysis in Section~\ref{sec:error_analysis}.

\section{Error Analysis}
\label{sec:error_analysis}

In this section, we present an in-depth analysis of the 
common error types exhibited by LLMs when summarizing CS 
dialogues. Our qualitative analysis across the EN-ZH, 
EN-TA, and EN-MS language pairs revealed three major 
failure modes:

\begin{itemize}
  \item \textbf{Code-Switching Loss (CSL)}: The summary 
primarily utilizes the English parts of the dialogue, 
ignoring or missing critical information from non-English 
segments.
  \item \textbf{Meaning Shift from Poor Translation (MST)}: 
The model misunderstands the code-switched segments, 
resulting in summaries that deviate from the dialogue's 
true meaning.
  \item \textbf{Speaker Misattribution (SMA)}: Summaries 
incorrectly assign statements to speakers, thus 
distorting intended meanings or misrepresenting 
participants' views.
\end{itemize}

\begin{table*}[t]
  \centering
  \small
  \begin{tabular}{l|l|cc|cc|cc}
  \toprule
  \textbf{Model} & \textbf{Lang} & \multicolumn{2}{c|}{\textbf{CSL}} & \multicolumn{2}{c|}{\textbf{MST}} & \multicolumn{2}{c}{\textbf{SMA}} \\
  & & Few-Shot & LoRA & Few-Shot & LoRA & Few-Shot & LoRA \\  
  \midrule
  
  \multirow{3}{*}{\textbf{Gemma-2-2B}} 
  & EN-ZH & 77.61 & 88.32 & 60.45 & 90.51 & 40.30 & 77.37 \\
  & EN-TA & 94.15 & 92.69 & 86.26 & 90.06 & 69.88 & 74.85 \\
  & EN-MS & 77.19 & 94.87 & 62.28 & 84.62 & 35.09 & 67.52 \\
  \midrule

  \multirow{3}{*}{\textbf{Qwen2.5-3B}} 
  & EN-ZH & 55.22 & 83.94 & 52.24 & 69.34 & 36.57 & 39.42 \\
  & EN-TA & 91.81 & 88.01 & 89.47 & 75.15 & 76.02 & 50.58 \\
  & EN-MS & 60.53 & 84.62 & 49.12 & 60.68 & 31.58 & 40.17 \\
  \midrule

  \multirow{3}{*}{\textbf{Qwen2.5-7B}} 
  & EN-ZH & 71.64 & 81.75 & 32.84 & 48.91 & 9.70 & 26.28 \\
  & EN-TA & 91.81 & 79.23 & 76.32 & 58.64 & 57.31 & 32.58 \\
  & EN-MS & 78.95 & 82.05 & 35.96 & 42.74 & 16.67 & 24.79 \\
  \midrule
  
  \multirow{3}{*}{\textbf{SEALLM-7B}}
  & EN-ZH & 64.18 & 84.67 & 41.79 & 70.80 & 21.64 & 33.58\\
  & EN-TA & 91.23 & 93.57 & 81.87 & 71.64 & 63.16 & 47.95\\
  & EN-MS & 68.42 & 85.47 & 45.61 & 69.23 & 23.68 & 40.17\\
  \midrule

  \multirow{3}{*}{\textbf{Mistral-7B}} 
  & EN-ZH & 55.97 & 83.21 & 37.31 & 69.34 & 22.39 & 40.88 \\
  & EN-TA & 88.30 & 89.47 & 81.58 & 79.82 & 66.37 & 53.22 \\
  & EN-MS & 67.54 & 77.78 & 44.74 & 70.09 & 28.07 & 44.44 \\
  \midrule

  \multirow{3}{*}{\textbf{Ministral-8B}} 
  & EN-ZH & 76.12 & 85.40 & 52.24 & 74.45 & 31.34 & 48.91 \\
  & EN-TA & 95.03 & 90.94 & 88.80 & 79.82 & 66.37 & 55.56 \\
  & EN-MS & 78.95 & 95.73 & 49.12 & 82.91 & 20.18 & 57.26 \\
  \midrule

  \multirow{3}{*}{\textbf{LLaMA-3-8B}} 
  & EN-ZH & 74.63 & 71.17 & 35.82 & 45.07 & 16.41 & 21.35 \\
  & EN-TA & 90.94 & \textbf{56.92} & 80.70 & 37.89 & 55.85 & 19.93 \\
  & EN-MS & 73.68 & 62.91 & 40.35 & 33.38 & 14.04 & 17.33 \\
  \midrule

  \multirow{3}{*}{\textbf{Gemma-2-9B}} 
  & EN-ZH & 59.70 & 85.40 & 18.66 & 57.66 &   8.21 & 35.77 \\
  & EN-TA & 71.35 & 82.75 & 35.09 & 67.84 &  22.22 & 46.20 \\
  & EN-MS & 59.65 & 79.60 & 18.42 & 52.53 &   8.77 & 29.45 \\
  \midrule

  \multirow{3}{*}{\textbf{SEA-LION-Gemma-2-9B}} 
  & EN-ZH & \textbf{53.73} & 83.94 & \textbf{11.19} & 76.64 &   \textbf{3.73} & 52.55 \\
  & EN-TA & 60.53 & 84.50 & \textbf{27.78} & 76.61 &  \textbf{17.54} & 49.71 \\
  & EN-MS & \textbf{48.25} & 86.32 & \textbf{14.04} & 64.96 &   \textbf{7.02} & 46.15 \\

  \bottomrule
  \end{tabular}
  \caption{Comparison of error rates (\%) in Few-Shot vs. LoRA fine-tuned models across different CS language pairs.}
  \label{tab:error_analysis}
\end{table*}

\begin{figure}
  \centering
  \includegraphics[width=\columnwidth]{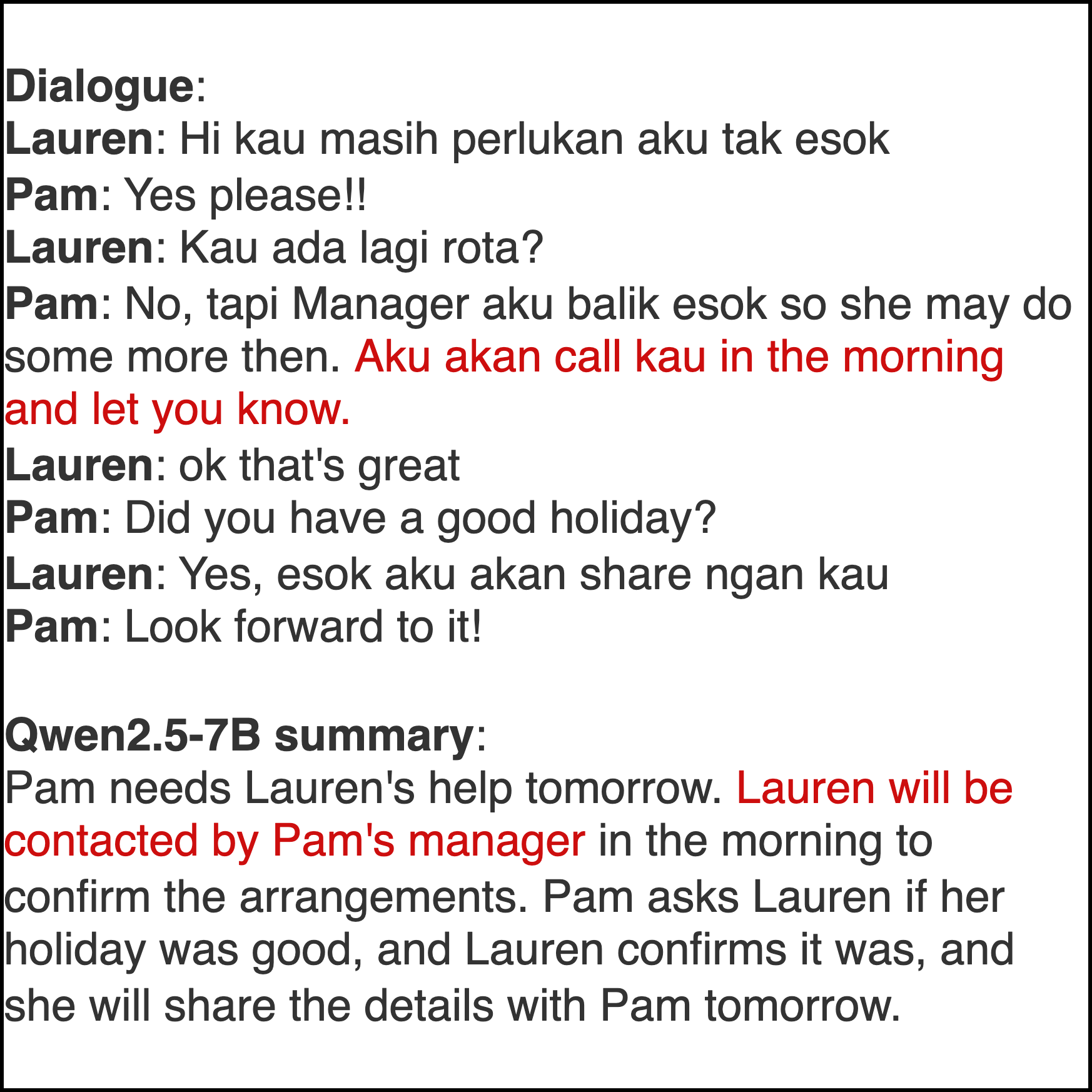}
  \caption{CSL error example}
  \label{fig:csl_error}
\end{figure}

Figure \ref{fig:csl_error} illustrates CSL, demonstrating 
how the summary incorrectly emphasizes English segments and 
misinterprets speaker intentions. Further examples across 
categories are provided in Tables \ref{tab:error_examples_1}, 
\ref{tab:error_examples_2}, and \ref{tab:error_examples_3}. 
Errors were identified automatically via GPT-4o, detailed 
further in Appendix \ref{appendix:error_analysis}.

To quantify these errors and compare model behaviors, we 
report the percentage of summaries exhibiting each error 
type across 9 LLMs under both Few-Shot and LoRA settings 
(Table \ref{tab:error_analysis}). The analysis reveals the 
following:

\paragraph{CSL persists as the dominant failure mode in Few-Shot evaluation.}
Across \textit{all} 9 models and the three language pairs, CSL exceeds 50\% 
in the Few-Shot setting (Table~\ref{tab:error_analysis}).  
Even the best performer, \textsc{SEA-LION-Gemma-2-9B}, attains a 
minimum of 53.73\% CSL on \textbf{EN-ZH}, confirming that LLMs 
systematically ignore non-English spans when summarizing 
code-switched dialogues, regardless of architecture or 
pre-training recipe.

\paragraph{Scaling within a model family does not reliably reduce errors.}
Comparing parameter-matched variants shows no monotonic gains: 
\textsc{Gemma-2-9B} lowers CSL on \textbf{EN-ZH} 
(77.61$\rightarrow$59.70) yet \emph{raises} Speaker Misattribution 
(SMA) on \textbf{EN-TA} by 7 pp over its 2B sibling, 
while \textsc{Qwen2.5-7B} improves CSL on \textbf{EN-MS} but 
leaves MST virtually unchanged.  These inconsistencies 
indicate that sheer scale is not a substitute for 
task-specific multilingual evaluation.

\paragraph{\textbf{EN-TA} exhibits consistently higher error rates across models.}
All evaluated models show their highest rates of CSL and SMA 
on EN-TA dialogues (e.g., \textsc{Qwen2.5-7B}: 91.81\% 
CSL, 57.31\% SMA). This suggests that certain language-specific 
properties, such as morphological complexity and syntactic 
divergence from English, may contribute to increased 
summarization difficulty in this setting.

\paragraph{SEA-oriented models excel in error analysis despite mediocre automatic scores.}
Regional models such as \textsc{SEALLM-7B} and \textsc{SEA-LION-Gemma-2-9B} 
achieve the lowest CSL and SMA within their parameter class 
(e.g., 53.73\% CSL on \textbf{EN-ZH}), yet trail larger English-specific models 
models on ROUGE and BERTScore.  This divergence reinforces that 
conventional metrics do not capture CS comprehension, 
underscoring the need for better metrics.

\paragraph{Synthetic fine-tuning amplifies errors under distribution shift.}
Fine-tuning on the Gemini-generated CS-Sum-Syn corpus 
degrades performance when the training distribution diverges 
from CS-Sum, most notably for \textbf{EN-ZH}, whose KL 
divergence is 2.48 versus $\le$0.55 for the other pairs 
(Table~\ref{tab:divergence}). After adaptation, 
\textsc{SEA-LION-Gemma-2-9B}’s CSL jumps from 53.73\% to 
83.94\% and MST from 11.19\% to 76.64\% 
(Table~\ref{tab:error_analysis}). These results illustrate 
that even synthetic data generated using a high-performing 
LLM does not improve the model’s ability to comprehend 
code-switched input, and the failure to produce data that 
enhances CS understanding also underscores the limitations 
of current LLMs comprehensibility of CS.

\section{Conclusion}
In this paper, we introduced CS-Sum, the first CS dialogue-to-English summarization 
benchmark across three language pairs: EN-TA, EN-ZH, and EN-MS. Our 
experiments revealed that LLMs perform inconsistently across CS pairs, 
with some open-source models outperforming SEA-specific models, highlighting 
that CS data must be present in pretraining, not just added through fine-tuning. 
Furthermore, our analysis identified three major error types—CSL, MST, and SMA—persisting 
across all models, demonstrating that current LLMs lack intrinsic CS comprehension. The 
quantitative results reinforce the severity of these failures, showing that fine-tuning 
alone does not fully resolve CS-specific challenges. To drive further research in multilingual 
NLP, we will release CS-Sum along with our codebase and evaluation framework.

\section{Limitations}
\label{sec:limitations}
While this work provides a comprehensive evaluation of CS dialogue 
summarization through the CS-Sum benchmark, there are some 
limitations that should be addressed in future work.

\begin{itemize}
    \item \textbf{Focus on Summarization:} Our experiments primarily 
focus on the task of summarization. However, code-switching also 
poses significant challenges for other long-context NLP tasks, such as 
machine translation, dialogue generation, and question answering. 
Extending the CS-Sum benchmark to these tasks would provide a broader 
understanding of LLMs' capabilities in handling CS data across 
different domains.
    
    \item \textbf{Finetuning with Synthetic Data:} The fine-tuning 
experiments in this paper are limited to synthetic code-switched data 
generated using Gemini-2. While this approach allowed us to assess the 
potential of fine-tuning for CS comprehension, it does not capture 
the complexities of real-world CS data. 

\end{itemize}

\bibliography{custom}

\appendix

\section{Synthetic Data Generation Process}
\label{app:synthetic_data}

To create a synthetic CS-dialogue summarization dataset, we combined the training sets of DialogSum and SAMSum, 
resulting in 19,014 English dialogue-summary pairs. We used Gemini-2-flash
to translate the English dialogues to code-switched dialogues in three CS 
pairs, as it is a large-scale SOTA LLM with a free-to-use API, making
it a practical choice for our experiments.

We used the following prompt to convert English dialogues into CS dialogues. Each prompt
was augmented with few-shot examples that were human generated.

\begin{tcolorbox}[colback=gray!10,colframe=gray!60,title=\textbf{Example Prompt for Malay-English Code-Switching}]
\small
You are a \{Malay/Tamil/Chinese\} person in your 20s.  

You are recruited for translating English dialogues to English-\{Malay/Tamil/Chinese\} code-switched dialogues.  

The code-switched dialogues should follow the same structure as the English dialogue.

This is for educational purposes, so DO NOT include swear words in your translation.  

Return just the code-switched dialogue.
\end{tcolorbox}

\section{Experiment settings}
\label{app:experiment_settings}
\subsection{Few-shot and Translate-Summarize}

The Few-Shot and Translate-Summarize settings evaluate the 
inherent CS comprehension of LLMs. Evaluating the LLMs in
this setting gives us an idea of their performance without
any additional training. 

\paragraph{Few-shot}
In the Few-Shot setup, we provide each LLM with three human-crafted examples containing code-switched dialogues and their corresponding summaries. 
The results from this setup, will quantify the inherent CS understanding of the LLMs for the three language pairs. 

\paragraph{Translate-Summarize}
In the Translate-Summarize approach, each CS dialogue is first translated into English before being summarized. This serves as a baseline to 
determine whether the presence of code-switching affects summary quality. The translation step is performed by the same LLM, ensuring that the 
summarization model receives only monolingual input. 

\subsection{LoRA and QLoRA}

In this subsection, we describe the fine-tuning process using \textit{LoRA} and \textit{QLoRA} to adapt open-source LLMs for CS dialogue 
summarization. Given that full fine-tuning is computationally expensive, we use parameter-efficient fine-tuning (PEFT) techniques to 
assess whether task-specific adaptation improves CS comprehension. The models are fine-tuned on CS-Sum-Syn.

\begin{table}[h]
    \centering
    \begin{tabular}{lccc}
        \hline
        \textbf{Model} & \textbf{r} & \textbf{$\alpha$}  & \textbf{LR} \\
        \hline
        Ministral-8B & 32 & 32 & 5e-5 \\
        LLaMA-3-8B & 16 & 16 & 5e-5 \\
        Qwen2.5-3B & 64 & 64 & 3e-5 \\
        Gemma-2-9B & 32 & 32 & 5e-5 \\
        SEA-Lion-Gemma-9B & 32 & 32 & 5e-5 \\
        Gemma-2-2B & 8 & 8 & 5e-5 \\
        Qwen-2.5-7B & 64 & 64 & 3e-5 \\
        Mistral-7B & 32 & 32 & 5e-5 \\
        SEALLM-7B & 64 & 64 & 3e-5 \\
        \hline
    \end{tabular}
    \caption{Hyperparameters used for LoRA and QLoRA fine-tuning.}
    \label{tab:lora_hyperparams}
\end{table}

The fine-tuning process follows model-specific hyperparameter configurations, as shown in Table~\ref{tab:lora_hyperparams}. 
The main hyperparameters include the rank parameter ($r$), scaling factor ($\alpha$), and learning rate ($lr$). These configurations were 
chosen based on empirical observations to balance training stability and performance.

All the models were trained for 4 epochs with a batch size of 8 and gradient 
accumulation for 8 steps. We followed the cosine learning rate scheduling after 
warming up for the first 3\% of steps. We used bf16 and gradient checkpointing
for memory efficient training. All the LLMs were trained on an H100 GPU and the
training lasted between 1 to 4 hours depending on the architecture and size of the 
LLM.

\clearpage

\section{Error Analysis Process}
\label{appendix:error_analysis}

The summaries generated by the LLMs typically exhibit the three major errors
listed in Section \ref{sec:error_analysis}. In this section, we describe the 
process we followed to detect these errors in the generated summaries.

We used GPT-4o to analyse the summary given the CS-dialogue for different errors.
For each error type, we used a separate prompt to guide the classification 
process. The LLM was asked to analyse the summary with respect to the particular
error type before providing a verdict on presence of the error. 

Below, we present the prompts used for each category.

\begin{tcolorbox}[colback=gray!10,colframe=gray!60,title=\textbf{Prompt for CSL}]
\small
You will be given a code-switched (CS) dialogue between two speakers and its generated summary. 
Identify whether the summary has ignored or removed key information from the non-English parts of the dialogue. 

The summary has an error if it meets one or more of the following criteria:
\begin{itemize}
    \item Does not include important content from the CS part of the dialogue.
    \item The summary only relies on the English part of the dialogue.
\end{itemize}

\end{tcolorbox}

\begin{tcolorbox}[colback=gray!10,colframe=gray!60,title=\textbf{Prompt for SMA}]
    \small
    You will be given a code-switched (CS) dialogue and its generated summary. 
    Analyze whether the summary has mixed up who said what.
    
    The summary has an error if it meets one or more of the following criteria:
    \begin{itemize}
        \item Has swapped roles or perspectives between speakers.
        \item Statements are wrongly assigned.
    \end{itemize}
    
\end{tcolorbox}

\begin{tcolorbox}[colback=gray!10,colframe=gray!60,title=\textbf{Prompt for MST}]
\small
You will be given a code-switched (CS) dialogue and its generated summary. 
Analyze whether the summary distorts the meaning of the original conversation due to incorrect understanding of the code-switching.

The summary has an error if it meets one or more of the following criteria:
\begin{itemize}
    \item The summary misinterprets the dialogue due to poor translation.
    \item Intent of the speakers has changed because of poor understanding of a CS phrase.
    \item Key points of the dialogue are misrepresented.
\end{itemize}

\end{tcolorbox}

\section{Training on filtered data}
\label{app:filtered_data}

\begin{figure}[t]
    \centering
    \includegraphics[width=\columnwidth]{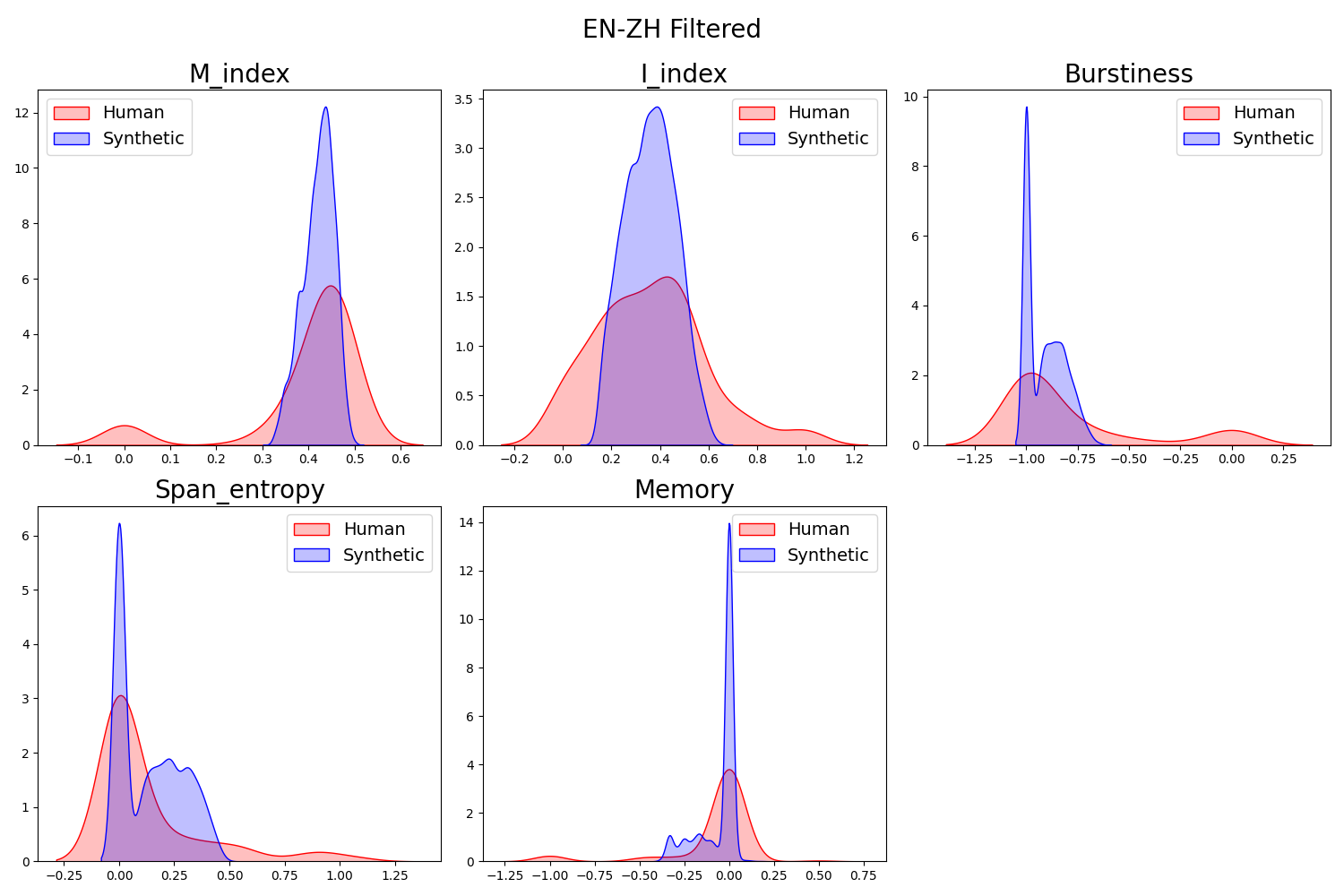}
    \caption{Distribution b/w filtered CS-Sum-Syn and CS-Sum for EN-ZH}
    \label{fig:filtered_data_zh}
\end{figure}

The difference in the distribution of CS-Sum and CS-Sum-Syn for 
EN-ZH (see Figure \ref{fig:en_zh}) is much more than the differences in distribution for EN-MS
and EN-TA. We hypothesize that this might be due to the tokenization
strategy of the LLMs which is more robust for Roman script compared to Chinese
script. The \% improvement gains for EN-ZH after LoRA and QLoRA finetuning
on EN-ZH is also less compared to the other two CS pairs (Table \ref{tab:lora_qlora_imp}).

To check if filtering CS-Sum-Syn to match CS-Sum's distribution would improve the 
finetuning results, we treat the CS metrics shown in Table \ref{tab:cs-metrics-cs-sum}
as a multivariate gaussian. We calculated the mean $ \boldsymbol{\mu} $ and the covariance 
matrix $ \Sigma $ of CS-Sum and used the Mahalanobis distance (eqn. \ref{eq:mahalanobis})
to calculate the distance between the instances in the training set and the test set. 

\begin{equation}
    D_M(\mathbf{x}) = \sqrt{ (\mathbf{x} - \boldsymbol{\mu})^T \Sigma^{-1} (\mathbf{x} - \boldsymbol{\mu}) }
    \label{eq:mahalanobis}
\end{equation}

We filtered the training data aggressively by selecting only the top 20 percentile of the 
training instances with the shortest distances. The filtered dataset had 3801 instances. 
The resulting distribution is shown in Figure \ref{fig:filtered_data_zh} which matches CS-Sum's
distribution better than the unfiltered one (see Figure \ref{fig:en_zh}).

\begin{figure*}[t]
    \centering
    \includegraphics[width=\textwidth]{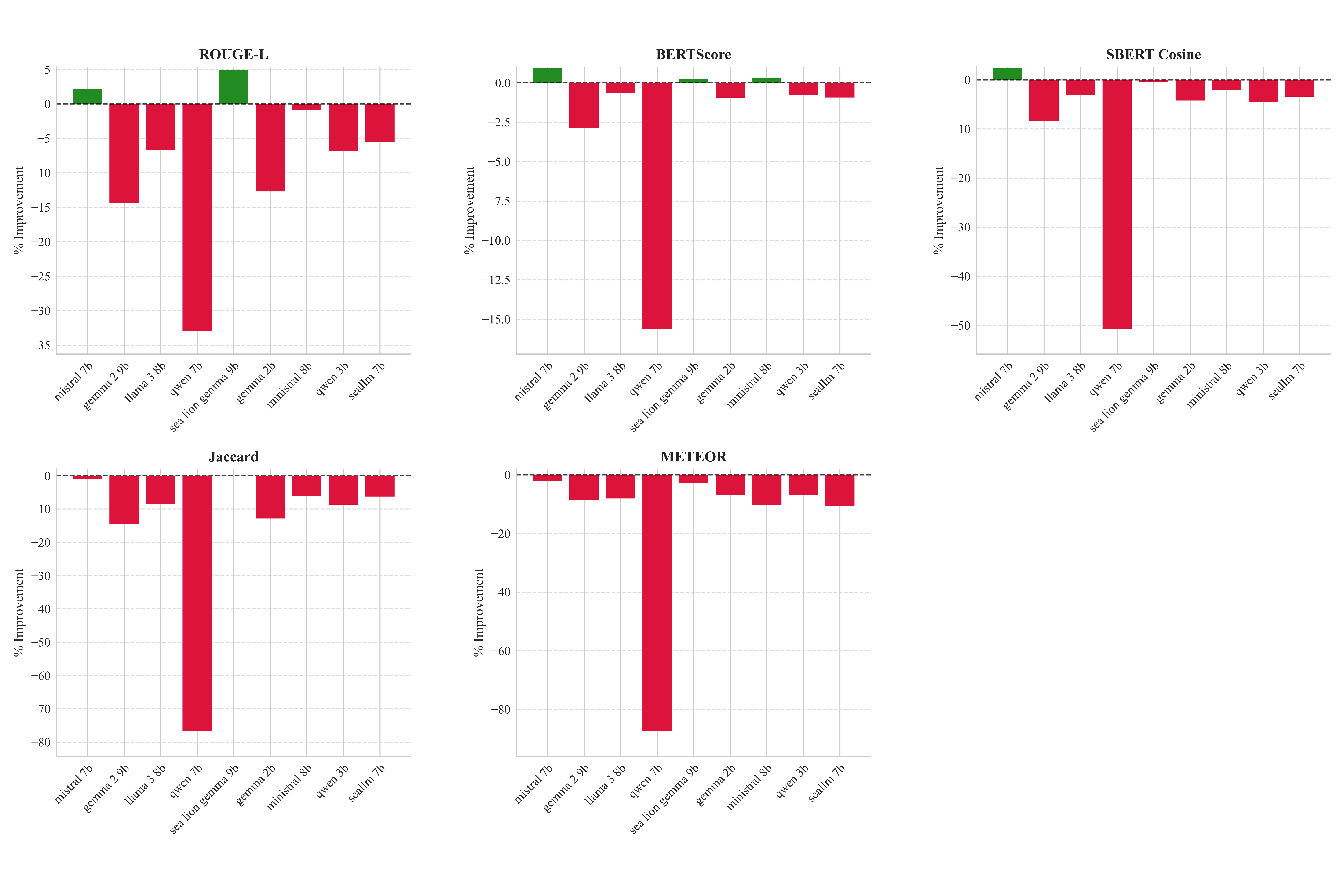}
    \caption{Filtered data \% improvement over CS-Sum-Syn for EN-ZH}
    \label{fig:filtered_data_imp}
\end{figure*}

We finetuned (QLoRA) all the 9 open-source LLMs on the filtered data for 3 epochs 
(4 for the unfiltered one) with all the other parameters staying the same. 
Figure \ref{fig:filtered_data_imp} shows the \% improvement of LLMs finetuned on
filtered data over the LLMs finetuned on the unfiltered one.

It can be seen that only 2 LLMs - Mistral 7B and SEA-Lion-Gemma-9B show 
actual improvement while other LLMs either don't improve much or have worse 
performances. Qwen2.5-7B suffers the worst when finetuned on filtered data.
This suggests that filtering using Mahalanobis distance may only be beneficial 
for certain architectures while being detrimental to others. Further investigation 
is required to determine if alternative filtering thresholds or strategies could 
yield better generalization across LLMs.

\clearpage

\begin{table*}[t]
  \centering
  % \small
  \resizebox{\textwidth}{!}{
  \begin{tabular}{l|l|ccccc}
  \toprule
  \textbf{Model} & \textbf{Lang} & \textbf{ROUGE-L} & \textbf{BERTScore} & \textbf{SBERT-COSINE} & \textbf{JACCARD} & \textbf{METEOR} \\
  \midrule

  \multirow{3}{*}{\textbf{Gemma-2-2B}} 
  & EN-ZH & \textcolor{green}{9.13} / \textcolor{green}{22.26} & \textcolor{green}{0.15} / \textcolor{green}{1.21} & \textcolor{green}{3.52} / \textcolor{green}{6.23} & \textcolor{green}{19.28} / \textcolor{green}{21.69} & \textcolor{green}{15.47} / \textcolor{green}{15.78} \\
  & EN-TA & \textcolor{green}{21.52} / \textcolor{green}{23.29} & \textcolor{green}{1.62} / \textcolor{green}{1.79} & \textcolor{green}{12.10} / \textcolor{green}{11.86} & \textcolor{green}{23.88} / \textcolor{green}{29.04} & \textcolor{green}{19.16} / \textcolor{green}{20.98} \\
  & EN-MS & \textcolor{green}{16.59} / \textcolor{green}{7.98} & \textcolor{green}{0.76} / \textcolor{green}{0.34} & \textcolor{green}{4.82} / \textcolor{green}{1.26} & \textcolor{green}{14.51} / \textcolor{green}{2.31} & \textcolor{green}{8.16} / \textcolor{red}{-4.92} \\
  \midrule

  \multirow{3}{*}{\textbf{Qwen2.5-3B}}
  & EN-ZH & \textcolor{green}{38.32} / \textcolor{green}{36.44} & \textcolor{green}{2.72} / \textcolor{green}{2.80} & \textcolor{green}{10.99} / \textcolor{green}{11.20} & \textcolor{green}{44.42} / \textcolor{green}{41.06} & \textcolor{green}{21.43} / \textcolor{green}{21.29} \\
  & EN-TA & \textcolor{green}{39.08} / \textcolor{green}{35.50} & \textcolor{green}{2.73} / \textcolor{green}{2.55} & \textcolor{green}{12.30} / \textcolor{green}{11.96} & \textcolor{green}{47.58} / \textcolor{green}{45.49} & \textcolor{green}{30.90} / \textcolor{green}{33.84} \\
  & EN-MS & \textcolor{green}{34.00} / \textcolor{green}{14.60} & \textcolor{green}{2.11} / \textcolor{green}{1.03} & \textcolor{green}{5.97} / \textcolor{green}{3.56} & \textcolor{green}{33.91} / \textcolor{green}{17.70} & \textcolor{green}{16.09} / \textcolor{green}{10.70} \\
  \midrule

  \multirow{3}{*}{\textbf{Qwen2.5-7B}} 
  & EN-ZH & \textcolor{green}{40.89} / \textcolor{green}{39.07} & \textcolor{green}{11.95} / \textcolor{green}{12.20} & \textcolor{green}{59.88} / \textcolor{green}{60.92} & \textcolor{green}{217.69} / \textcolor{green}{208.36} & \textcolor{green}{234.54} / \textcolor{green}{239.63} \\
  & EN-TA & \textcolor{green}{22.86} / \textcolor{green}{20.85} & \textcolor{green}{1.83} / \textcolor{green}{1.72} & \textcolor{green}{11.50} / \textcolor{green}{11.10} & \textcolor{green}{32.72} / \textcolor{green}{30.17} & \textcolor{green}{35.66} / \textcolor{green}{36.26} \\
  & EN-MS & \textcolor{green}{9.71} / \textcolor{green}{2.06} & \textcolor{green}{0.74} / \textcolor{red}{-0.10} & \textcolor{green}{6.27} / \textcolor{green}{3.64} & \textcolor{green}{12.16} / \textcolor{green}{1.82} & \textcolor{green}{11.38} / \textcolor{green}{5.36} \\
  \midrule

  \multirow{3}{*}{\textbf{Mistral-7B}} 
  & EN-ZH & \textcolor{green}{17.91} / \textcolor{green}{2.18} & \textcolor{green}{0.73} / \textcolor{red}{-0.80} & \textcolor{green}{7.78} / \textcolor{green}{2.43} & \textcolor{green}{28.10} / \textcolor{green}{15.19} & \textcolor{green}{20.10} / \textcolor{green}{9.41} \\
  & EN-TA & \textcolor{red}{-1.79} / \textcolor{red}{-3.62} & \textcolor{red}{-2.20} / \textcolor{red}{-2.85} & \textcolor{green}{3.90} / \textcolor{green}{1.90} & \textcolor{green}{8.93} / \textcolor{green}{6.68} & \textcolor{green}{13.76} / \textcolor{green}{11.95} \\
  & EN-MS & \textcolor{red}{-11.88} / \textcolor{green}{15.12} & \textcolor{red}{-2.93} / \textcolor{green}{0.56} & \textcolor{red}{-1.21} / \textcolor{green}{2.93} & \textcolor{red}{-8.56} / \textcolor{green}{8.96} & \textcolor{green}{3.83} / \textcolor{red}{-2.09} \\
  \midrule

  \multirow{3}{*}{\textbf{Ministral-8B}} 
  & EN-ZH & \textcolor{green}{9.92} / \textcolor{green}{13.16} & \textcolor{green}{1.15} / \textcolor{green}{0.69} & \textcolor{green}{9.58} / \textcolor{green}{9.85} & \textcolor{green}{30.37} / \textcolor{green}{34.98} & \textcolor{green}{29.41} / \textcolor{green}{33.61} \\
  & EN-TA & \textcolor{green}{18.38} / \textcolor{green}{20.88} & \textcolor{green}{1.41} / \textcolor{green}{1.41} & \textcolor{green}{13.73} / \textcolor{green}{14.13} & \textcolor{green}{34.19} / \textcolor{green}{36.83} & \textcolor{green}{35.43} / \textcolor{green}{39.83} \\
  & EN-MS & \textcolor{red}{-24.12} / \textcolor{green}{7.51} & \textcolor{red}{-2.73} / \textcolor{green}{0.63} & \textcolor{red}{-1.64} / \textcolor{green}{4.51} & \textcolor{red}{-8.07} / \textcolor{green}{9.89} & \textcolor{green}{7.89} / \textcolor{green}{2.76} \\

  \bottomrule
  \end{tabular}
  }
  \caption{LoRA / QLoRA \% Improvement over Few-Shot on CS-Sum}
  \label{tab:lora_qlora_imp_2}
\end{table*}

\clearpage

\begin{table*}[h!]
    \centering
    \resizebox{\textwidth}{!}{
      \begin{tabular}{|p{5cm}|p{5cm}|p{3cm}|p{3cm}|}
        \toprule
        \textbf{EN Dialogue} & \textbf{CS Dialogue} & \textbf{LLM Summary} & \textbf{Error Explanation} \\
        \midrule
        Matthew: I have a cold. Need y'all support. It's painful \newline
        Lisa: Fight it. Drink orange juice. Make some soup \newline
        Matthew: I'll try \newline
        Lisa: I believe in u \newline
        ... &
        Matthew: Enakku cold irukku. Unga elllaroda support um vennum. Athu romba painful \newline
        Lisa: Atha fight pannu. Orange juice kudi. Soup pannu \newline
        Matthew: \textcolor{red}{Naa try panren} \newline
        Lisa: Naa unna believe panran \newline
        ... & 
  Matthew is feeling unwell due to a cold and an allergy. \newline
  Lisa suggests that he drink orange juice and soup to feel better. \newline 
  \textcolor{red}{Matthew tries it and feels a bit better.} & 
  Matthew says that he will try the pill but the
  summary mentions that he had tried it and feels better \\
  \midrule
  Anna: is anyone going to pick Mark from the airport? \newline
  Marcus: i could but when and where from? \newline
  Anna: Sydney, Thursday at 3 \newline
  Marcus: am or pm? :D \newline
  Leslie: haha fortunately pm:D \newline
  Marcus: hmm i have a meeting at 1. I don't think i can make it \newline
  Leslie: well i guess it will take him some time after landing, reclaiming luggage etc \newline
  Anna: yeah I reckon it's fine if you're there at 4 \newline
  Marcus: oh well ok then \newline
  ... & 
  \begin{CJK*}{UTF8}{gbsn}
  Anna：有人去机场接 Mark 吗？\newline
  Marcus：我可以，但是什么时候去哪里呢？\newline
  Anna：Sydney，星期四 3 点 \newline
  Marcus：am 还是 pm？:D \newline
  Leslie：haha，幸运的是下午 :D \newline
  Marcus：hmm 我 1 点有个会议。我想我去不了了 \newline
  Leslie：well 我想 after his landing 他需要一些时间，reclaiming luggage etc \newline
  Anna：\textcolor{red}{是的，我认为如果你 4 点到那里没问题了} \newline
  Marcus：oh 那好吧 \newline
  ... \newline
  \end{CJK*} & 
  Marcus originally plans to pick up Mark at the airport but has a meeting at 1 PM 
  and can't make it. \textcolor{red}{Leslie suggests Marcus arrive after 4 PM} to 
  give Mark enough time to reclaim luggage. Anna will inform 
  Mark about Marcus's new plan. & 
  Anna was the one who suggest Marcus to go
  at 4 PM which is wrongly attributed to Leslie \\
  \midrule
  
  Ken: You are an idiot. \newline
  Greg: What? \newline
  Ken: Yeah you are, I want my money back \newline
  Greg: Was your account hacked by some prankster? \newline
  Ken: No, I'm the prankster, just having a laugh at your expense \newline
  Greg: Well, you are an idiot too then XD & 
  \begin{CJK*}{UTF8}{gbsn}
  Ken：操你，你 pimp \newline
  Greg：什么？\newline
  Ken：操你，我要回我的钱 \newline
  Greg：你的账户被恶作剧者 hacked 了吗？\newline
  Ken：\textcolor{red}{不，我才是恶作剧者，只是开个玩笑以你为代价} \newline
  Greg：\textcolor{red}{Well，那你也操 XD}
  \end{CJK*} & 
  Ken and Greg were joking with each other.
  Ken played a prank on Greg \textcolor{red}{but now he wants his money back.} 
  Greg was not happy about it & 
  The dialogue is just a friendly banter and
  there’s actually no transaction of money. \\
  \bottomrule
  
      \end{tabular}
    }
    \caption{Error examples}
    \label{tab:error_examples_1}
  \end{table*}
  
  \begin{table*}
    \centering
    \resizebox{\textwidth}{!}{
      \begin{tabular}{|p{5cm}|p{5cm}|p{3cm}|p{3cm}|}
        \toprule
        \textbf{EN Dialogue} & \textbf{CS Dialogue} & \textbf{LLM Summary} & \textbf{Error Explanation} \\
        \midrule 
  ... \newline
  Person2: Large deposits of coal were found nearly and so many industries located themselves here. The village quickly grew into a key industrial centre. \newline
  Person1: As the city grew, it must have absorbed many village nearby. \newline
  Person2: Yes, it did. The names of those village survive as the names of parts of the city. \newline
  Person1: I see. Are there any building more than 200 years old in your city? \newline
  Person2: Oh, yes. Several of the buildings from the villages still survive. Many of them were inns for travelers and today survive as pubs. There was a castle near one village, so our city has a castle too. \newline
  Person1: Really? So your city does have some old history after all. &
  \begin{CJK*}{UTF8}{gbsn}
  ... \newline
  Person2: 附近发现了大量 coal. 许多工业都设在这里。这个村庄迅速发展成为一个重要的industrial centre。\newline
  Person1: 随着城市发展，它一定absorbed了附近的许多村庄。\newline
  Person2: 是的。这些村庄的名字作为城市部分地区的名字而留存下来。\newline
  Person1: I see. 你的城市里有超过 200 年历史的building 吗？\newline
  Person2: 村庄里的一些buildings仍然保存完好。其中许多都是旅客的inns，如今作为pubs保留了下来。一\textcolor{red}{个村庄附近有一座castle，所以我们的城市也有一座castle。} \newline
  Person1: 真的吗？原来你们城市也有悠久的 history 啊。
  \end{CJK*}&
  Person2 tells Person1 how their city grew and has some 
  buildings over 200 years old. \textcolor{red}{Person2 also introduces a 
  castle nearby} &
  The summary is based on only the English
  words of the dialogue and also Person2 talks about a
  castle, does not introduce it \\
  \midrule
  Alan: <file\_photo> \newline
  Alan: look what I just found \newline 
  Robert: dude, that's just nasty and you know it  \newline 
  Robert: it has no sugar, no taste, and additional cinnamon flavoring \newline
  Alan: yeah, I know - that's awesome  \newline
  Robert: you sir have a very strange tastes :P \newline
  Alan: well, and I found a perfect company for it <file\_photo> \newline
  Robert: oh, that's more like it! \newline
  Robert: but does the whiskey go well with the cinnamon? flavored whiskey is the worst... \newline
  Alan: Actually it does taste surprisingly well. The cinnamon is not overpowering. If you put enough whiskey that is  \newline
  ... \newline 
  & 
  Alan: <file\_photo> \newline
  Alan: tengok ape aku jumpe  \newline
  Robert: dude, that's just nasty and you know it  \newline
  Robert: it has no sugar, no taste, and additional cinnamon flavoring \newline
  Alan: yeah, I know - itulah yang awesome  \newline
  Robert: you sir ada tastes pelik :P \newline
  Alan: well, and \textcolor{red}{I found a perfect company for it} <file\_photo> \newline
  Robert: oh, that's more like it! \newline
  Robert: Tapi does the whiskey go well with the cinnamon? flavored whiskey rabak... \newline 
  Alan: Actually it does taste surprisingly well. The cinnamon tak terlalu kuat. If you put enough whiskey that is  \newline
  ...  \newline
  & 
  Alan shares a photo of a cinnamon-flavored whiskey with Robert, 
  who initially finds it disgusting. Alan defends the unique 
  taste and \textcolor{red}{shares another photo of him enjoying it with a 
  companion}. Robert jokingly points out Alan's unusual taste 
  preferences but admits the cinnamon taste isn't overpowering 
  when paired with enough whiskey. They reminisce about a 
  previous cherry flavor and agree it was better than the 
  current sugar-free version. 
  & 
  The summary wrongly states that Alan has
  a companion when he was actually referring to eating
  cinnamon along with it \\
  \bottomrule 
      \end{tabular}
    }
    \caption{Error examples}
    \label{tab:error_examples_2}
  \end{table*}
  
  \begin{table*}
    \centering
    \resizebox{\textwidth}{!}{
      \begin{tabular}{|p{5cm}|p{5cm}|p{3cm}|p{3cm}|}
        \toprule
        \textbf{EN Dialogue} & \textbf{CS Dialogue} & \textbf{LLM Summary} & \textbf{Error Explanation} \\
        \midrule
  Callan: Something's wrong with my Samsung S8. \newline
  Wade: What, exactly? \newline
  Callan: All I get is a black screen. \newline
  Wade: Did you drop it or something? \newline
  Callan: I think I know, this has happened before. \newline
  Wade:  It did? What happened before? \newline
  Callan: It overheats and then it takes a while for it to restart again.. \newline
  Wade: That sucks.. Be carefull with the battery, it can be dangerous. \newline
  ...
  & 
  Callan: Samsung S8 aku ada problem. \newline
  Wade: What, exactly? \newline
  Callan: Aku asyik dapat black screen. \newline
  Wade: Kau terjatuhkan ke apa? \newline
  Callan: Aku rasa aku tahu kot, ni pernah happened kat aku. \newline
  Wade:  It did? What happened before? \newline
  Callan: \textcolor{red}{Dia overheat pastu it takes a while for it to restart again..} \newline
  Wade: That sucks.. Be carefull with the battery, boleh jadi bahaya. \newline
  ...
  &
  The speaker, Callan, is having issues with his Samsung S8 as it keeps 
  getting black screens. \textcolor{red}{Wade suggests it might be 
  overheating}. Callan intends to take it to a store as it's still 
  under warranty
  &
  Callan was the one who suggested the phone
  might be overheating \\
  \midrule
  Person1: What do you know about our company? \newline
  Person2: Well, as I know this company is one of the largest and best in this field of business. It mainly produces toys for children. It employs more than 10, 000 people throughout the world. The president now is Mr. Jackson. The Shanghai branch was founded five years ago with a staff of more than 2, 000. \newline
  Person1: Very good. If you enter this company, what department would you like to work in? \newline
  Person2: May I work in sales department? \newline
  Person1: OK. Oh, how much do you make at your current job, if you don't mind my asking? \newline
  Person2: I'm paid 2, 000 yuan per month plus some allowances and bonus as that come to about 3, 000 yuan a month. How much will you pay me? \newline
  Person1: Well, the starting salary for a clerk in our company is 2, 500 yuan monthly and after three months he would get a raise if his work is satisfactory. \newline
  Person2: Do you have any fringe benefits? \newline
  ...
  & 
  Person1: Enga company pathi enna therium ungaluku? \newline
  Person2: \textcolor{red}{Indha field indha company dhaan largest and best nu therium}. Main focus vandhu children toys produce panradhu. Current president Mr. Jackson. Shanghai la 5 year munnadi oru branch start panirkeenga. Anga 2000 employees irukanga. Total ah 10000 employees. \newline
  Person1: Very good. Inga job kedaicha, endha department la work pananum ungaluku? \newline
  \textcolor{red}{Person2: Sales department la?} \newline
  Person1: OK. Current job la evlo earn panreenga? If you don't mind my asking? \newline
  Person2: Current salary 2000 yuan plus konjam allowances and bonus ku apram 3000 yuan per month. Neenga evlo pay panuveenga? \newline
  Person1: Clerks ku starting salary 2500 yuan. 3 months ku apram raise varum. \newline
  Person2: Fringe benefits edhavadhu iruka? \newline
  ...
  &
  \textcolor{red}{Person2 is describing their current company} which is the largest and 
  best in the field, focusing on children's toys.  They have a branch 
  in Shanghai with 2000 employees, totaling 10,000 employees worldwide. 
  \textcolor{red}{Person1 is interested in a sales position and asks about the salary 
  and benefits}, learning that Person2 currently earns 3000 yuan per 
  month with allowances and bonuses. Person1 explains that the starting 
  salary for clerks is 2500 yuan, with a raise after three months,
  plus benefits like semi-annual bonuses, a small Spring Festival bonus, 
  and 4 weeks of
  &
  LLM has completely switched Person1 and
  Person2 in the first half of the summary \\
  \bottomrule
      \end{tabular}
    }
  \caption{Error examples}
  \label{tab:error_examples_3}
  \end{table*}

\end{document}